\pdfoutput=1

\documentclass[11pt]{article}

\usepackage[final]{acl}

\usepackage{times}
\usepackage{latexsym}
\usepackage{tikz}
\usepackage{amssymb}
\usepackage{subcaption}
\usepackage{makecell}

\usepackage[T1]{fontenc}

\usepackage[utf8]{inputenc}

\usepackage{microtype}

\usepackage{inconsolata}

\usepackage{graphicx}
\usepackage{pifont}
\usepackage{booktabs}       
\usepackage{url}
\usepackage{enumitem} 
\usepackage{xcolor,colortbl}

\title{\textsc{AssistantBench}: Can Web Agents Solve \\ Realistic and Time-Consuming Tasks?}

\author{\makecell{Ori Yoran$^{1}$~~~~~ Samuel Joseph Amouyal$^{1}$~~~~~ Chaitanya Malaviya$^{2}$ \\~~~~~ Ben Bogin$^{3,4}$ ~~~~~ Ofir Press$^{5}$ ~~~~~ Jonathan Berant$^{1}$ } \\ 
$^{1}$Tel Aviv University\hspace{5mm}
$^{2}$University of Pennsylvania\hspace{5mm} \\ $^{3}$Allen Institute for AI \hspace{5mm} 
$^{4}$University of Washington \hspace{5mm} $^{5}$Princeton University\\ 
\texttt{\small\makecell{\{ori.yoran, samuel.amouyal, joberant\}@cs.tau.ac.il}}}

\newif\ifcomments
\commentstrue
\ifcomments
    \newcommand\oy[1]{\textcolor{red}{[OY: #1]}}
    \newcommand\cm[1]{\textcolor{magenta}{[CM: #1]}}
    \newcommand\jb[1]{\textcolor{blue}{[JB: #1]}}
    \newcommand\sa[1]{\textcolor{brown}{[SA: #1]}}
    \newcommand\ofir[1]{\textcolor{purple}{[ofir: #1]}}
    
    \newcommand\sandy[1]{\textcolor{orange}{[Sandy: #1]}}

\else
    \providecommand{\oy}[1]{}
    \providecommand{\tw}[1]{}
    \providecommand{\jb}[1]{}
    \providecommand{\orir}[1]{}
    \providecommand{\ofir}[1]{}
    \providecommand{\sandy}[1]{}
    \providecommand{\sa}[1]{}
    \providecommand{\cm}[1]{}

\fi

\begin{document}
\maketitle

\newcommand{\numtasks}{$214$}
\newcommand{\numparticipants}{$53$}
\newcommand{\numgeneralparticipants}{$18$}

\newcommand{\numexpertparticipants}{$35$}
\newcommand{\numuniquepages}{$525$}
\newcommand{\numuniquedomains}{$258$}

\newcommand{\numfanouttasks}{$31$}
\newcommand{\assistbench}{\textsc{AssistantBench}}
\newcommand{\seeact}{\textsc{SeeAct}}
\newcommand{\seeactexplorer}{\textsc{SPA}}
\newcommand{\fanoutqa}{\textsc{FanoutQA}}
\newcommand{\wob}{\textsc{MiniWob++}}
\newcommand{\fone}{F$_1$}
\newcommand\google{\textsc{Google Search}}
\newcommand\cbinst{\textsc{CB-Inst}}
\newcommand\cbones{\textsc{CB-1S}}
\newcommand\ralminst{\textsc{RALM-Inst}}
\newcommand\ralmones{\textsc{RALM-1S}}
\newcommand\gaia{\textsc{GAIA}}
\newcommand\webshop{\textsc{WebShop}}
\newcommand\webarena{\textsc{WebArena}}
\newcommand\visualwebarena{\textsc{VisualWebArena}}
\newcommand\mindtoweb{\textsc{Mind2Web}}
\newcommand\webvoyager{\textsc{WebVoyager}}
\newcommand\mmina{\textsc{MMInA}}
\newcommand\chatgpt{\textsc{ChatGPT}}
\newcommand\ralmonesfallback{\ralmones{}$\rightarrow$\textsc{CB}}
\newcommand\ralminstfallback{\ralminst{}$\rightarrow$\textsc{CB}}
\newcommand\seeactfallback{\seeact{}$\rightarrow$\textsc{CB}}
\newcommand\spafallback{\seeactexplorer{}$\rightarrow$\textsc{CB}}

\begin{abstract}
Language agents, built on top of language models (LMs), are systems that can interact with complex environments, such as the open web.
In this work, we examine whether such agents can perform realistic and time-consuming tasks on the web, e.g., monitoring real-estate markets or locating relevant nearby businesses.
We introduce \assistbench{}, a challenging new benchmark consisting of \numtasks{} realistic tasks that can be automatically evaluated, covering different scenarios and domains. We find that \assistbench{} exposes the limitations of current systems, including language models and retrieval-augmented language models, as no model reaches an accuracy of more than $26$ points. While closed-book LMs perform well in terms of accuracy, they exhibit low precision and tend to hallucinate facts. State-of-the-art web agents reach a score of near zero. Additionally, we introduce \textsc{SeePlanAct} (\seeactexplorer{}), a new web agent that significantly outperforms previous agents, and an ensemble of \seeactexplorer{} and closed-book models reaches the best overall performance. Moreover, we analyze failures of current systems and highlight that open web navigation remains a major challenge.\footnote{Code, data, leaderboard, and models are available at \url{https://assistantbench.github.io}.}

\end{abstract}

\section{Introduction}

\begin{figure}[t]
  \includegraphics[width=1.0\columnwidth]{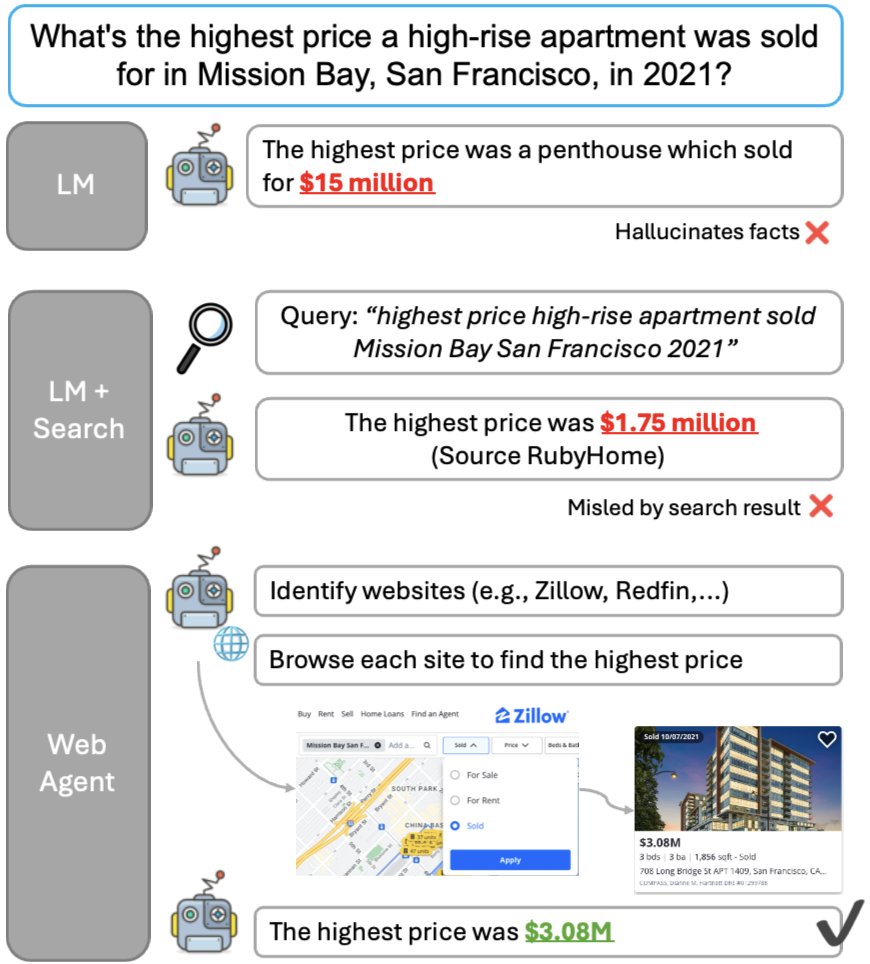}  
  \caption{An example task in \assistbench{}. 
  Current LMs and retrieval-augmented LMs are limited in their ability to accomplish such tasks compared to an agent that can interact with the web.
  We propose \assistbench{}, which includes \numtasks{} diverse tasks, in addition to a new agent, and evaluate our agent as well as other existing systems on \assistbench{}.
  }
  ~\label{intro_fig}
  \vspace{-10pt}
\end{figure}

\begin{figure}[t]
  \includegraphics[width=1.0\columnwidth]{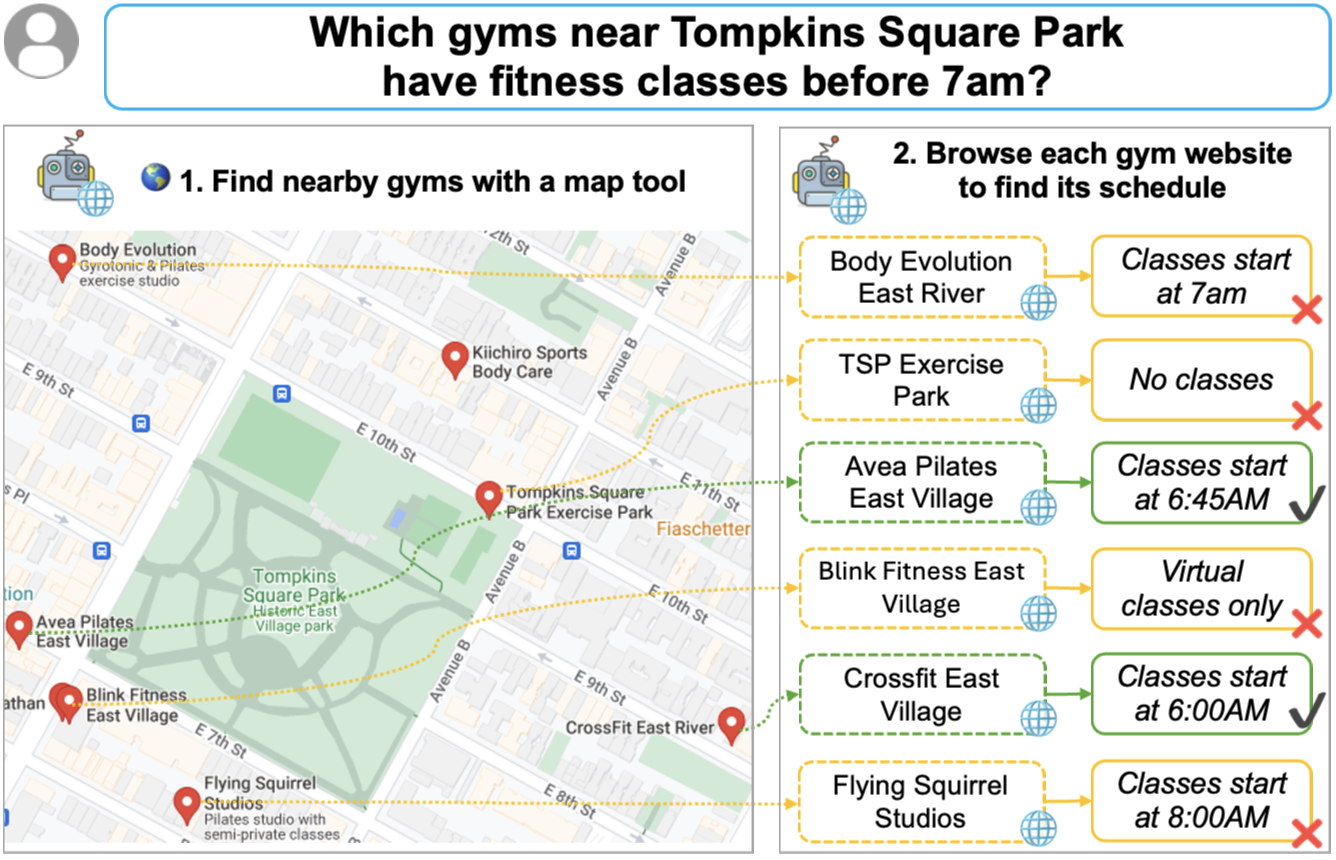}  
  \caption{A gold trajectory for a task in \assistbench{}. A web agent can solve the task by first interacting with a map tool (e.g., Google Maps) to find nearby gyms and then browsing each website to find the relevant schedule.
  }
  ~\label{fig:trajectory}
  \vspace{-10pt}
\end{figure}

\begin{figure*}[t]
  \includegraphics[width=\textwidth]{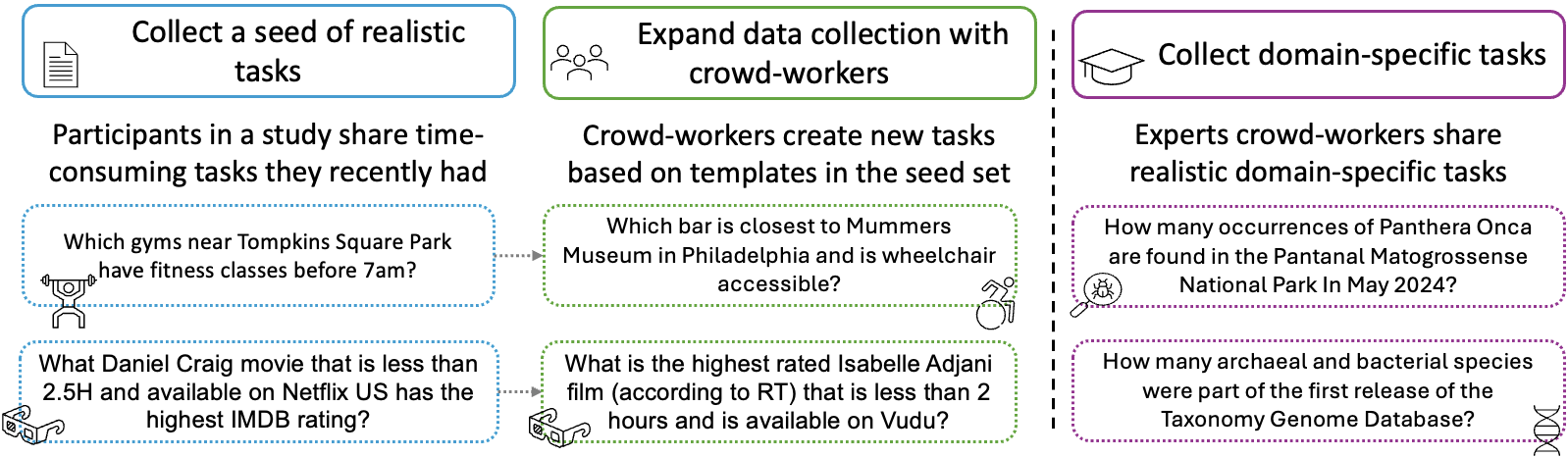}  
\caption{The main steps in our data collection pipeline. (Left) Participants in our study share time-consuming tasks they recently performed. (Center) We expand the dataset by showing tasks as templates to crowd-workers and ask them to create similar tasks. (
Right)  To increase the diversity of tasks to additional domains, we collect domain-specific tasks with domain-expert crowd-workers. }

  ~\label{fig:datacollection}
  \vspace{-10pt}
\end{figure*}

Consider a person looking to buy a house who is monitoring the current real-estate market (Fig.~\ref{intro_fig}), or a fitness enthusiast on vacation in New York looking for an early-bird fitness class (Fig.~\ref{fig:trajectory}).
Automated systems for such information-seeking tasks have the potential to greatly improve information access for users. However, current models are limited in the way they gather information for completing such user requests, making them unable to answer most of these queries. 
Assisting users in these scenarios is challenging for models that rely solely on parametric knowledge \cite{roberts-etal-2020-much}, as they cannot readily access information from the web and are prone to hallucinations \cite{xiao2021hallucination, lin2022truthfulqa} (Fig.~\ref{intro_fig}, top).
Retrieving relevant evidence can help \cite{realm,lewis2020rag,ram2023incontext,10.5555/3648699.3648950}, but benefits are limited by the quality of evidence retrieved. Furthermore, retrieving irrelevant evidence can sometimes even hurt performance \cite{yoran2024making} (Fig.~\ref{intro_fig}, center).
A more promising approach is to simulate what humans do -- an AI system could search the web to find relevant web pages, interact with them and synthesize the information gathered to produce an output 
\cite{nakano2022webgpt, deng2023mindweb, zhou2024webarena} (Fig.~\ref{intro_fig}, bottom).

Web agents, powered by language models, present an opportunity to assist users with time-consuming web tasks. Such tasks are commonplace, yet agents are currently not evaluated in the full scope of their actual usage. Existing benchmarks proposed to evaluate web agents focus on tasks that require interacting with a single website \cite{liu2018reinforcement, yao2022webshop, deng2023mindweb, he2024webvoyager} or tasks that require operating within a sandbox environment \cite{zhou2024webarena, koh2024visualwebarena} (see \S\ref{sec:rw} for a detailed review of related works).
While these benchmarks have been crucial for building better performing web agents \cite{zheng2024seeact, he2024webvoyager}, they are not sufficient for evaluating the ability of agents to plan and reason over the entire web \cite{lecun2022path}.

In this work, we present \assistbench{}, a new benchmark with \numtasks{} diverse tasks, aimed to evaluate the ability of web agents to browse the entire web and solve real-world tasks that are time-consuming for humans.
Tasks in \assistbench{} are based on real information needs encountered by humans. To solve these tasks, an agent must autonomously browse the web to identify relevant web pages and dynamically interact with them to produce an output (Fig. \ref{fig:trajectory}).

To create \assistbench{}, we first asked
\numgeneralparticipants{} participants to share recent information-seeking tasks that they could solve using the web but required a few minutes of browsing. We then expanded this set by asking crowdworkers to use tasks from this seed set as templates for new tasks. To increase the diversity of \assistbench{} to professional assistance tasks, we also asked expert crowdworkers to share recent tasks which required expertise in their field and were time-consuming.
At the end of this process, we collected \numtasks{} tasks from $53$ different people, \numexpertparticipants{} of them domain experts, that require browsing more than \numuniquepages{} web pages from \numuniquedomains{} different websites and span a wide range of topics.

To study the abilities of current models to solve tasks in \assistbench{}, we evaluate strong closed-book and retrieval-augmented models, as well as \seeact{} \cite{zheng2024seeact}, a state-of-the-art web agent. In addition, we introduce \textsc{SeePlanAct} (\seeactexplorer{}), a variant of \seeact{} equipped with a planning component and a memory buffer (\S\ref{sec:model}). We find that \assistbench{} is challenging for all models, with no model reaching accuracy of more than $26$ points (\S\ref{sec:experiments}).
Our proposed agent, \seeactexplorer{}, outperforms \seeact{} by about $7$ points, answering twice as many questions with a precision that is $5$ points higher. Finally, an ensemble that combines \seeactexplorer{} with a closed-book model achieves the best overall performance.

We analyze why \assistbench{} is challenging for current systems (\S\ref{sec:analysis}) and
find that tasks provided by experts are most challenging. We observe that trajectories that are very short or very long often lead to errors.
We then conduct a detailed qualitative analysis, showing that errors during web navigation, such as choosing an incorrect trajectory or getting stuck in a loop are most frequent for web agents ($60\%$ and $37\%$ of errors for \seeact{} and \seeactexplorer{}, respectively) and that
closed-book models often generate hallucinated facts ($85\%$ of errors). Finally, failures in retrieving relevant information are the most common failure mode for retrieval-augmented models ($80\%$ of errors). Moreover, we find that proprietary chatbots such as ChatGPT suffer from similar problems.

To conclude, our main contributions are:
\begin{itemize}[itemsep=0.0pt,topsep=0.2pt]
  \item We release \assistbench{}, a new benchmark for web agents that contains \numtasks{} realistic and time-consuming web tasks.
  \item We propose \seeactexplorer{}, a web agent equipped with memory and planning components for multi-hop, info-seeking questions. 
  \item We show that \assistbench{} is challenging for current systems (closed-book models, retreival-augmented and agents), with the best model reaching an accuracy of $26$ points.
\end{itemize}

\section{Background}
\label{sec:background}
The importance of web agents in assisting with daily tasks has led to growing interest in their evaluation.
Earlier works introduced environments that model web interaction as a reinforcement learning task \cite{pmlr-v70-shi17a, liu2018reinforcement}.
Recent benchmarks propose simplified or static environments that are more realistic and support a diverse set of tasks \cite{yao2022webshop, deng2023mindweb, koh2024visualwebarena, he2024webvoyager, zhang2024mmina}.
Recently, \citet{zhang2024mmina} proposed \mmina{}, a dataset for multi-hop web tasks over $14$ websites.
Most similar to our work, is \gaia{} \cite{mialon2024gaia} a benchmark to evaluate AI assistants with challenging tasks that require tools, including open-web browsing. However, many tasks in GAIA also require video or audio processing tools,\footnote{Even the filtered set of $90$ \gaia{} web tasks used in \webvoyager{} \cite{he2024webvoyager} includes tasks that require video processing, such as: \emph{`In the video} \url{https://www.youtube.com/watch?v=L1vXCYZAYYM}\emph{, what is the highest number of bird species to be on camera simultaneously?'}} while our focus is solely on open-web navigation. 

\assistbench{} differs from previous benchmarks by focusing on time-consuming tasks, such as the example in Fig.~\ref{fig:trajectory} (see Tab.~\ref{tab:benchmarks_examples} in \S\ref{appendix_comparison} for an example task from each benchmark), featuring tasks that are more realistic, require multiple websites, cover a wide range of domains, and are extremely challenging for current models. Hence, we are hopeful it can provide a  good measuring stick for the ability of web agents to assist humans.


\section{\assistbench{}}
\label{sec:dataset}

We now describe the process of constructing \assistbench{}. 
We first explain the criteria for our tasks, followed by our data collection pipeline, and conclude with an analysis of the collected tasks.

\subsection{Criteria for \assistbench{} Tasks}
\label{dataset_criteria}

Our focus is on realistic and challenging tasks that can be solved with the web, without using additional tools (e.g., video processing) and satisfy the following criteria:

\begin{itemize}[itemsep=0.0pt,topsep=0.2pt]
  \item \textbf{Realistic}: A task that is likely to answer a real human need.
  \item \textbf{Time-consuming}: A challenging task that takes a person at least several minutes to perform.
  \item \textbf{Automatically verifiable}: A task with a closed-form answer that can be automatically verified and is unlikely to change quickly.\footnote{We remove tasks whose answers are likely to change in the near future and manually verify each task by two authors. We further discuss time-dependency in \S\ref{sec:time_dependency}.}
\end{itemize}

\subsection{Data Collection}
Building \assistbench{} involved three main steps: (a) creating a seed set of tasks, (b) expanding the dataset with crowd-workers, and (c) collecting tasks with domain experts (see Fig.~\ref{fig:datacollection}).

\paragraph{Seed tasks.}
In this step (Fig.~\ref{fig:datacollection}, left), we asked
participants in a study to share time-consuming web browsing tasks that they recently had to perform. Due to the complexity of annotation and to ensure efficient communication, we limited participation to individuals with whom we had direct contact in addition to this paper's authors. We provide the instructions shown in \S\ref{appendix_seed}, Fig.~\ref{fig:survey_instructions}.

We made several design choices to verify collected tasks fit all our criteria. 
Each task was manually reviewed and potentially tweaked when necessary (e.g., adding a date constraint so the task is not time-dependent). In addition to the task, we also collected the 1) gold answer, 2) URLs where the answer can be verified, and 3) an explanation for how the task can be solved. Overall, we collected $72$ tasks from $18$ participants. We note that coming up with diverse tasks was challenging for participants, which results in each participant providing no more than a handful of tasks. We provide examples and additional statistics in \S\ref{appendix_seed}.

\paragraph{Expanding \assistbench{}.}
We used tasks collected in the seed set as templates for new tasks. Consider the task: \emph{What Daniel Craig movie that is less than 150 minutes and available on Netflix US has the highest IMDB rating?} shown in  Fig.~\ref{fig:datacollection} (left). Similar tasks can be created by changing the actor name, runtime, or rating system for this task (Fig.~\ref{fig:datacollection}, center). To increase the number of tasks in \assistbench{}, we presented 
crowdworkers with tasks from the seed set and asked them to use these as templates to create new tasks.
As with the seed tasks, we collected the gold answers, URLs, and explanations, and manually validated each task.
We collected an additional $102$ tasks using this method. See \S\ref{appendix_expanding_set} for additional statistics and technical details about the data collection task.

\paragraph{Domain-specific tasks.}
To increase the diversity and skill level of tasks in \assistbench{}, we asked domain experts to share realistic web-solvable tasks they performed in their professional lives.
We recruited $35$ participants through Prolific who came from a range of fields including biology, geography, visual arts, etc. Additional details about the participants are provided in \S\ref{appendix_expert_collection}.
First, we ran a qualification task where we asked participants to share professional websites they often use, e.g., online historical archives or analysis tools. Then, we filtered out websites that were unlikely to meet our data collection criteria, e.g., websites that require logins or involve scant interactions. We then asked experts to create tasks that require interaction with these websites.
We manually verified each example in this set and made small refinements, e.g., adding a date constraint when needed. We verified updates to tasks with the original participants to ensure the final tasks reflected the domain experts' needs. Overall, our expert set contains $42$ tasks, including two tasks from the seed set that require high domain-expertise.

\subsection{Data Statistics}
\paragraph{Task distribution.} In total, we collected \numtasks{} unique tasks covering different users, domains, and websites: $70$ tasks from $18$ users in the general seed set, which were expanded to $172$ tasks with the help of crowdworkers, and $42$ tasks in the expert set. Tasks require different answering strategies, as exemplified in Fig.~\ref{intro_fig} and Fig.~\ref{fig:trajectory}, and cover diverse domains -- the expert set itself spans more than $15$ different domains (e.g., Biology, Law, Medicine). Moreover, tasks require interacting with many websites; the answers are spread across \numuniquepages{} webpages on \numuniquedomains{} websites.

\paragraph{Development set.} We put the majority of tasks in the test set. To facilitate experimentation and analysis, we create a development set with $33$ tasks ($12$ from the seed set and $21$ from crowd-workers). We provide more examples from this set in \S\ref{appendix_dev_set}.

\subsection{Automatic Evaluation}
\label{subsec:eval}
To automatically evaluate answers, we support three answer types: strings, numbers, and dictionaries. We allow tasks with up to five answers (see 
\S\ref{appendix_dev_set} for examples). For strings, we use \fone{} between the predicted and gold answer words. We also support list answers using the official implementation from \citet{dua-etal-2019-drop}, which aligns predicted and gold answers based on their similarity. For numbers, to let models get partial credit when close to the answer, we use a metric similar to the order magnitude metric from \citet{kalyan-etal-2021-much}, but give a score of zero once the prediction $A'$ is an order of magnitude from the gold answer $A$, i.e.,  $\max \{0, 1-\log\frac{\max(A, A')}{\min(A, A')}$\}.

Some tasks in \assistbench{}, such as finding tuition fees for daycare centers, require a more structured output since we need to evaluate the correct value for each entity (e.g., the center and its tuition fee). To support semi-structured outputs, we provide the model with the output format in the task input (e.g., \emph{The answer should be a json with the keys `center' and `price (USD)'}). 
We then compare the two dictionaries by looking at the matching values of identical keys and use our evaluation methods based on the value's type (i.e., we use our word-level \fone{} evaluation when it is a string and the order of magnitude evaluation when it is a number).
When one of the keys is missing from the other dictionary, the model receives a score of zero for the matching key-value pair.  We calculate recall (i.e., the average score for gold key-value pairs) and precision (i.e., the average score for predicted key-value pairs) and return the \fone{} score, inspired by recent metrics in tabular question-answering \cite{liu-etal-2023-deplot}.

In some cases, models abstain from generating an answer. Hence, we also measure the \textit{answer rate} (fraction of tasks for which an answer was generated) and \textit{precision} (the average accuracy when the model did not abstain).
We also report \emph{exact match}, where the model gets $1$ if it generated the exact reference answer, and $0$ otherwise.

\section{\seeactexplorer{}: See-\textit{Plan}-Act}
\label{sec:model}

We now describe \seeactexplorer{}, a web agent built to solve tasks in \assistbench{}. We begin by describing \seeact{} \cite{zheng2024seeact}, a state-of-the-art web agent, which \seeactexplorer{} is built upon.

\begin{figure}[t]
  \includegraphics[width=1.0\columnwidth]{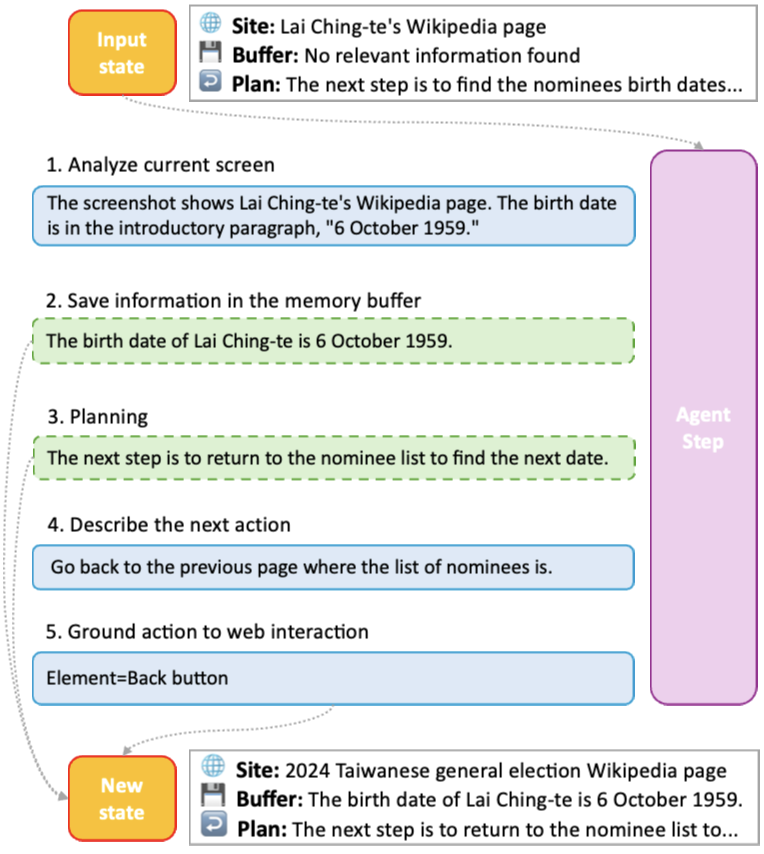}  
  \caption{An example execution step for \seeactexplorer{} (the full trajectory is shown in \S\ref{sec:appendix_fanout}, Fig.~\ref{fig:fanout_success}). The model's state includes the current website, the memory buffer, and the execution plan. At each step, the model can add information to the memory buffer and refine its plan (green with dashed border lines). The model generates the next action by taking a screenshot of the current page, describing the action, and grounding it to the HTML, similar to \seeact{} \cite{zheng2024seeact} (blue).}
  ~\label{fig:agent}
  \vspace{-10pt}
\end{figure}

\paragraph{\seeact{}.} \seeact{} \cite{zheng2024seeact} is a multi-modal web agent that looks at web pages and interacts with their HTML elements.
Each step has two main stages: (1)  the agent examines a screenshot that explains what the next action should be in natural language, and (2) the explanation is grounded to an action with one of the HTML elements. Fig.~\ref{fig:agent} presents an overview of a step in \seeact{} and its differences from \seeactexplorer{}.

\paragraph{\seeactexplorer{}.} Since \assistbench{} focuses on tasks that require \textit{planning and reasoning} \cite{lecun2022path}, we equip \seeact{} with two specialized components: (1) a planning component for the model to plan and re-plan its execution, and (2) a memory component with the option to transfer information between steps via a memory buffer.
Fig.~\ref{fig:agent} shows an example of a full step of \seeactexplorer{}, including the planning and memory component (in green), which are implemented through prompting.
In addition, to support open-web navigation, we equip the agent with new actions that enable (a) returning to a previous page, (b) navigating to a specified URL, or (c) entering a query directly into a search engine. We provide additional technical details in \S\ref{appendix_models}.

\section{Experiments}
\label{sec:experiments}

We now describe our experimental setup. 
First, we describe the models we evaluate.
Then, we describe \fanoutqa{}, an  additional benchmark we use to assess the ability of models to tackle information-seeking tasks on a single website, Wikipedia. Finally, we present results for all models.

\begin{table*}[t]
\centering
\footnotesize
  \begin{tabular}{llcccc}
    \toprule
    Type & Model  & Acc. & Ans. $\%$ & Prec. & EM
    \\ \midrule 
    Closed-book LMs& \cbinst{} & 16.5 & 53.6 & 30.7 & 6.1 \\
    & \cbones{} & 22.2 & 89.5 &  24.8 & 8.3 \\ \midrule
    Retrieval-augmented LMs & \ralminst{} & 11.8 & 60.2 & 19.5 & 5.5 \\
    & \ralmones{} & 10.7 & 48.1 & 22.4 & 3.9 \\ \midrule
   Web agents & \seeact{} & 4.1 & 15.5 & 26.3 & 2.2 \\ 
    & \seeactexplorer{} (ours) & 11.1 & 35.9 & \textbf{30.9} & 5.5\\ \midrule
    Ensembles & \ralminstfallback{} & 18.7 & \textbf{93.9} & 19.9 & 6.6 \\ 
      & \ralmonesfallback{} & 19.5 & 92.8 & 21.0 & 6.1  \\ 
     & \seeactfallback{} & 23.4 & 89.5 & 26.1 & 9.4 \\ 
     & \spafallback{} (ours) & \textbf{25.2} & 91.7 & 27.5 & \textbf{9.9} \\ 
  \bottomrule
\end{tabular}
\caption{Results on the \assistbench{} test set with GPT4-T.}
  \label{tab:test_results}
\end{table*}

\subsection{Models}
\label{subsec:models}
In addition to \seeact{} and \seeactexplorer{}, we evaluate four additional strong baselines on \assistbench{}; two LMs and two retrieval-augmented LMs:

\begin{itemize}[itemsep=0.0pt,topsep=0.2pt]

\item{\textbf{Closed-book, instruction-tuned (\cbinst{})}}: a zero-shot model prompted to answer tasks using chain-of-thought prompting \cite{wei2022chain, nye2022show}. 
We experiment with GPT-4-Turbo~\cite{openai2024gpt4} and Claude-3.5-Sonnet \cite{anthropic2024claude} as our closed-book models.

\item \textbf{{Closed-book, one-shot (\cbones{}):}} we add one in-context example. Specifically, the model generates a chain-of-thought as a series of intermediate questions and answers (i.e., self-ask prompting  \cite{press-etal-2023-measuring}, see \S\ref{appendix_prompts} for all prompts).

\item \textbf{{Retrieval-augmented, instruction-tuned 
(\ralminst{}):}} we prompt the model to use a search engine as a tool, similar to ReAct \cite{yao2023react}. We use \google{} as our retriever.

\item \textbf{{Retrieval-augmented, one-shot (\ralmones):}} we add one example of self-ask prompting with search where retrieval is called for each intermediate question.

\end{itemize}

\paragraph{Ensembles.}

As we will show, web agents often refrain from answering when web navigation fails. Thus, we evaluate ensembles where we fall back to closed-book models, \cbones{}, when the agents abstain. For example, we denote by \ralmonesfallback{} the model that falls back to \cbones{} when \ralminst{} abstains.

\paragraph{Implementation details of web agents.} 
Following previous work \cite{zheng2024seeact}, we use GPT-4-Turbo~\cite{openai2024gpt4} and Claude-3.5-Sonnet \cite{anthropic2024claude} as the base model for our retrieval-augmented LMs and web agents\footnote{Experiments with open-weight Llama-3.2 models are in-progress and results will be added to the leaderboard.}. To avoid infinite loops and following recent work showing that agents succeed quickly and fail slowly \cite{yang2024sweagent}, we limit the number of execution steps of our agents to $30$. We provide all technical details in \S\ref{appendix_models} and our prompts in \S\ref{appendix_prompts}.

\subsection{Info-seeking Tasks over Wikipedia}
\label{subsec:fanout}
Tasks in \assistbench{} require interacting with multiple websites. \fanoutqa{} \cite{zhu2024fanoutqa} is a benchmark that includes tasks that require aggregating information from multiple web pages (e.g., \emph{What is the total number of employees in the five largest banks in the world?}) but answers to all questions can be found on Wikipedia. 
Hence, we use \fanoutqa{} as an additional development set to evaluate the benefits of our agent, \seeactexplorer{}.

We make several design choices to transfer tasks from \fanoutqa{} to our setting: (1) we focus on tasks that have dictionary answers (\S\ref{subsec:eval}), and (2) we keep only tasks where our prompted closed-book model (\cbones{}) does not perfectly answer the question (see \S\ref{sec:appendix_fanout} for additional details). At the end of this process, we are left with \numfanouttasks{} tasks on which we evaluate all models.

\subsection{Results}
\label{results}

We now present results for all systems on \assistbench{}, as well as \fanoutqa{}.

\paragraph{\assistbench{}.}

 Tab.~\ref{tab:test_results} presents results on the \assistbench{} test set with GPT4-T. 
 All systems perform poorly, with no system reaching more than $25\%$ accuracy.
 Our agent, \seeactexplorer{}, outperforms \seeact{} by $20\%$ in answer rate and $5$ precision points, and has the best precision.
 Interestingly, closed-book models (\cbinst{}, \cbones{}) have better accuracy than retrieval-augmented LMs and web agents, mainly due to their higher answer rate. Adding an in-context example increases accuracy and answer rate, at the cost of a drop in precision. An ensemble that falls backs from \seeactexplorer{} to \cbones{} when the agent abstains has the best overall accuracy, albeit a slightly lower precision than \seeactexplorer{} and \cbinst{}. When using the more strict exact match, we observe similar trends, and no model obtains more than $9.9\%$ points.

We present results with Claude-3.5-Sonnet in \S\ref{appendix_results}, Tab.~\ref{tab:test_results_claude}. We observe similar trends with \spafallback{} achieving the best accuracy of $26.4$ and \seeactexplorer{} outperforming \seeact{} by $10$ points with better precision and answer rates.
Additionally, we present results on our development set in \S\ref{appendix_results}, Tab.~\ref{tab:dev_results}, with similar trends.
 
To assist future work in focusing on tasks that are behind the reach of current closed-book models we 
annotate each task with a  difficulty level based on the accuracy of closed-book models: \emph{easy} tasks are cases where both GPT-4 and Claude-3.5-Sonnet \cbones{} achieve an accuracy of $0.5$ or higher, \emph{medium} tasks are when one of the closed-book models' accuracy is equal or larger than $0.5$, and \emph{hard} tasks are cases none of the two answers correctly. We present results for the different difficulty levels in \S\ref{appendix_results}, Tab.~\ref{tab:difficulty_levels}. The majority of tasks are hard for all models, and \seeactexplorer{} significantly outperforms \seeact{} on medium and hard tasks.

\begin{table}[t]
\centering
\footnotesize
  \begin{tabular}{lcccc}
    \toprule
    Model  & Acc. & Ans. $\%$ & Prec. & EM
    \\ \midrule 
    \cbinst{} & 34.8 & 93.5 & 37.2  & 0.0 \\
    \cbones{} & \textbf{40.9} & \textbf{100.0} & 40.9 & 0.0 \\
    \ralminst{} & 9.6 & 93.5 & 10.2 & 0.0 \\
    \ralmones{} & 27.3 & 93.5 & 29.2 & 0.0 \\
    \seeact{} & 7.5 & 16.1 & 46.4 & 0.0 \\ \midrule
    \seeactexplorer{} (ours) & 30.0 & 61.3 & \textbf{48.9} & \textbf{9.7} \\
  \bottomrule
\end{tabular}
\caption{Results on \fanoutqa{} with GPT4-T.}
  \label{tab:fanout}
\end{table}

\paragraph{\fanoutqa{}.}
Tab.~\ref{tab:fanout} presents results on \fanoutqa{}, where we observe similar trends. \seeactexplorer{} outperforms \seeact{} by $22.5$ points and has an answer rate that is higher by $45\%$. Interestingly, \seeactexplorer{} has a higher or similar precision relative to all other models. In addition, it succeeds in reaching the exact reference answer in $3/31$ tasks, more than any other model (see \S\ref{sec:appendix_fanout}, Fig.~\ref{fig:fanout_success} for an example). 
Closed-book models were used to filter examples and have an exact match score of zero by design. However, despite their lower precision, they still perform best in terms of overall accuracy.

\begin{table}[t]
\centering
\footnotesize
  \begin{tabular}{lccc}
    \toprule
    Model  & Accuracy & Accuracy & Accuracy \\
      & (seed) & (expanded) & (experts)
    \\ \midrule 
    \cbinst{}  & 15.8 & \textbf{19.6} & 11.3\\
    \cbones{} & \textbf{29.7} & 18.7 & \textbf{19.5}\\
    \ralminst{} & 13.4 & 12.7 & 9.7\\
    \ralmones{}  & 7.9 & 13.3 & 11.4\\
    \seeact{} & 2.9 & 1.9 & 9.7 \\ \midrule
    \seeactexplorer{} & 12.9 & 11.8 & 7.6 \\
  \bottomrule
\end{tabular}
\caption{Accuracy for the different \assistbench{} data collection methods with GPT4-T.}
  \label{tab:expert_split}
\end{table}

\section{Analysis}
\label{sec:analysis}
 \begin{figure}[t]
  \includegraphics[width=1.0\columnwidth]{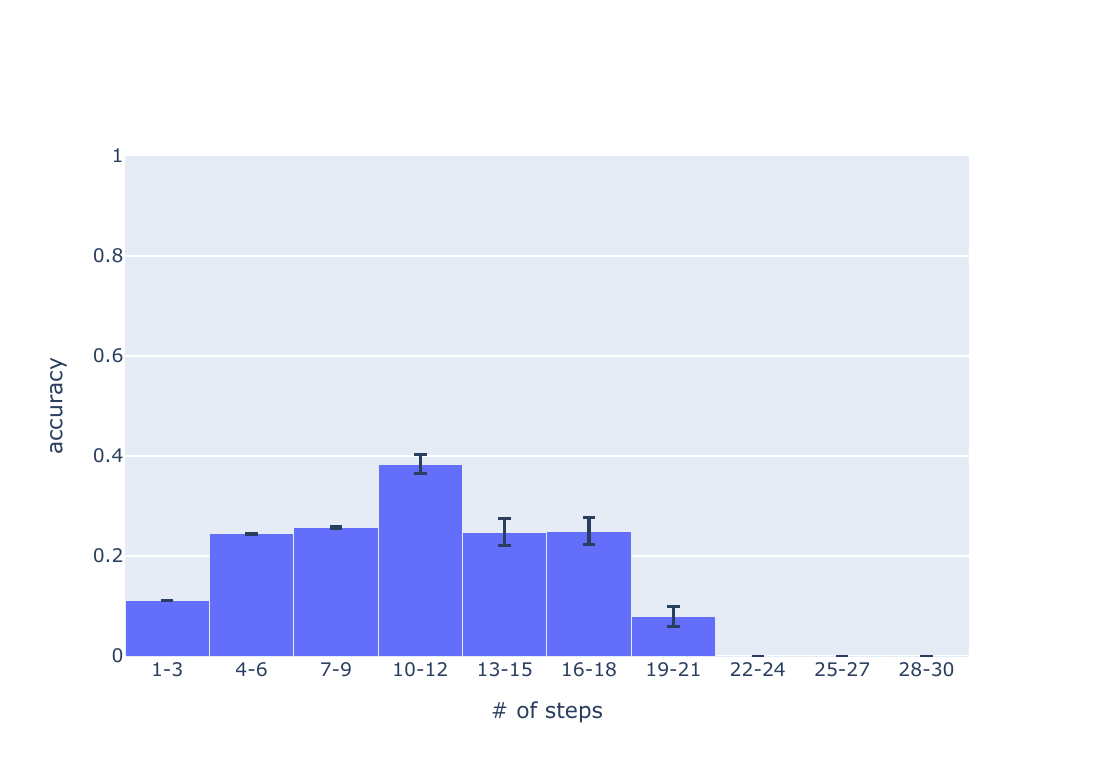}  
  \caption{Accuracy on the \assistbench{} test set for \seeactexplorer{} as a function of the number of execution steps. Error bars indicate standard error of the mean. Web agents struggle when trajectories are very long or very short, and peak at trajectories of around ten steps.}
  ~\label{fig:seeact_num_calls}
  \vspace{-10pt}
\end{figure}

\begin{figure*}[t]
  \includegraphics[width=\textwidth]{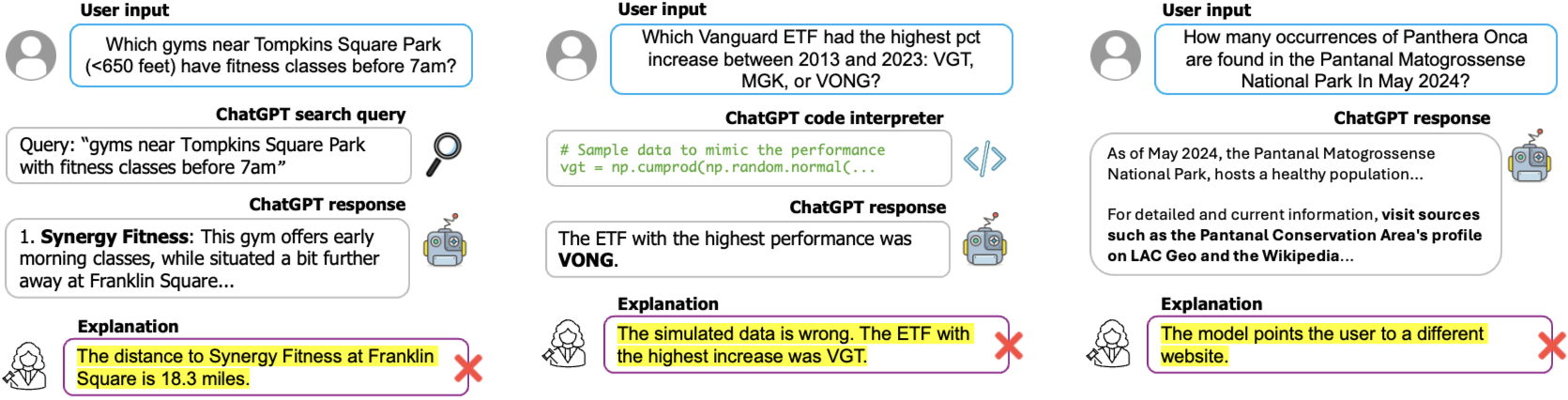}  
\caption{Failure cases for \chatgpt{}. Tasks are presented at the top, above \chatgpt{} generations and an explanation for each phenomenon. The most common failure is for the model to over-rely on search results and generate a wrong answer (left). In some cases, the model hallucinates non-factual information in the code interpreter which leads to wrong answers (center, the code generation is not directly shown to the user). Rarely, the model abstains from answering and points the user to a different website (right).}

  ~\label{fig:chatgpt_analysis}
  \vspace{-10pt}
\end{figure*}
To understand the challenges in solving \assistbench{}, we  analyze performance across different data collection strategies, followed by a qualitative analysis of model errors. Then, we examine how popular chat-bots respond to tasks from \assistbench{}.

\subsection{When is \assistbench{} Challenging?}
\label{sec:quantitative_analysis}

\paragraph{Expert tasks introduce different challenges.} 
Tab.~\ref{tab:expert_split} presents accuracy results for our different data collection strategies. 
The expert set is more challenging for \cbinst{} but is actually easier for \seeact{} than the general set. This is likely due to the fact that the majority of answers in the expert set can be found in a single URL ($70\%$) versus only $20\%$ for the general set. We present all other evaluation metrics for each strategy in \S\ref{appendix_results}, Tab.~\ref{tab:full_results_splits}. 

\paragraph{Web agents fail fast or slow.}

Both \seeact{} and \seeactexplorer{} get near zero accuracy when trajectories are very long (more than $15$ actions). Accuracy is also near zero when trajectories are very short (less than $5$ steps), and peaks at around ten steps (see Fig.~\ref{fig:seeact_num_calls} for \seeactexplorer{} and \ref{appendix_analysis}, Fig.~\ref{fig:sa_num_calls} for \seeact{}). As tasks in \assistbench{} require long web interactions, our results can explain why they are challenging for current agents.
Interestingly, we identify a similar trend with our multi-step retrieval-augmented LMs (\ref{appendix_analysis}, Fig.~\ref{fig:num_steps}).

\subsection{Why is \assistbench{} Challenging for Current Systems?}
\label{sec:error_analysis}

Next, we manually analyze errors for our models on the \assistbench{} development set.

\paragraph{\textsc{Web Agents}.} The majority of \seeactexplorer{} and \seeact{} errors are due to the model not providing an answer ($80\%$ and $97\%$ of errors respectively). We identify four error cases for not generating an answer and present their frequency in Tab.~\ref{tab:error_analysis}: (1) \emph{Navigation errors} (the model's trajectory is wrong and will not reach the gold answer, see \S\ref{appendix_analysis}, Fig.~\ref{fig:navigation_failure} for an example) (2) \emph{Grounding} (the agent fails to interact with the page properly, e.g, fails to click on an element, or does not recognize elements on the page) (3) \emph{No answer} (the model sees the correct answer but does not return it), and (4) \emph{Technical issues} (the agent crashes due a technical issue; see \S\ref{appendix_models} for more technical details). The majority of errors, $36.7\%$ and $63.6\%$ for \seeactexplorer{} and \seeact{} respectively, are due to \emph{navigation errors}, highlighting that solving tasks in \assistbench{} is challenging for current agents. On rare cases, the models reach the correct answer but do not generate it. Similar to the original \seeact{} work \cite{zheng2024seeact}, we find that \emph{grounding} is a major challenge causing around $20\%$ of errors. The stronger \seeactexplorer{} has higher answer rates but that also leads to more wrong answers.

\begin{table}[t]
\centering
\footnotesize
  \begin{tabular}{lcc}
    \toprule
    Cause  & \seeactexplorer{} & \seeact{}\\ \midrule
    Navigation error & $36.7\%$ & $63.6\%$ \\
    Grounding & $23.3\%$ & $18.2\%$ \\
    Technical issue & $16.7\%$ & $9.1\%$ \\
    No answer & $3.3\%$ & $6.1\%$\\ \midrule
    Wrong answer & $20.0\%$ & $3.0\%$ \\

  \bottomrule
\end{tabular}
\caption{Error analysis for \seeactexplorer{} and \seeact{}. For both agents, most errors are due to failures to generate an answer, with \emph{navigation error} the most prominent cause.}
  \label{tab:error_analysis}
\end{table}

\paragraph{Closed-book Models.}
When closed-book models abstain from answering, they often suggest a plan for how the task can be solved (around $90\%$ of abstains). When closed-book models give an incorrect answer, they tend to hallucinate answers ($85\%$ of errors) or provide outdated answers ($15\%$ of errors). We provide examples in \S\ref{appendix_analysis}, Fig.~\ref{fig:closed_book_failures}.

\paragraph{Retrieval-augmented Models.}

Most errors are caused by failure to retrieve relevant information ($80\%$ of errors). We categorize retrieval failures to three common classes: (a) tool-related (the information can be found using a tool on the web, e.g., distances with a map), which occurs in $38.5\%$ of cases, (b) partial information: the retriever returns only partial information from the gold web page, usually when it needs to retrieve large amounts of data ($11.5\%$ of cases), and (c) irrelevant retrieved context ($50\%$ of the cases). Our analysis shows that tasks in \assistbench{} are challenging and beyond the reach of current retrieval-augmented systems.
The other $20\%$ of the errors are due to the model failing to use relevant retrieval results or exploring trajectories that do not lead to the gold answer. We provide examples in \S\ref{appendix_analysis}, Fig.~\ref{fig:ralm_failures}.

\subsection{Can \chatgpt{} with Web Search solve \assistbench{}?} 
\label{sec:chatgpt_analysis}

To evaluate the ability of proprietary chatbots to solve tasks from \assistbench{}, we use \chatgpt{} with tasks from our development set.\footnote{\url{https://chatgpt.com}: we use GPT-o with web search and code execution, with the temporary chat option.} Similar to our models, \chatgpt{} errs on more than $90\%$ of the tasks. We identify several common failures (Fig.~\ref{fig:chatgpt_analysis}). The majority of errors are due to the model over-relying on search results to generate wrong answers (Fig.~\ref{fig:chatgpt_analysis}, left). In another common failure (about $15\%$ of the cases, Fig.~\ref{fig:chatgpt_analysis}, center), the model hallucinates non-factual information in the code interpreter which is then presented to the user. On rare cases, the model is able to abstain by pointing the user to relevant websites (Fig.~\ref{fig:chatgpt_analysis}, right). Our results suggest that tasks in \assistbench{} expose the limitations of current systems that do not navigate the web.

\subsection{Time-dependency of tasks in \assistbench{}}
\label{sec:time_dependency}
When collecting examples (\S\ref{sec:dataset}), we focus on tasks whose answers are unlikely to change in the near future. To analyze time-dependency of tasks in \assistbench{}, we categorize tasks to three classes: (a) static tasks that have a date constraint and will not change with time, e.g., \emph{Which Vanguard ETF had the highest pct increase between 2013 and 2023}, accounting for the $43.5\%$ of tasks, (b) tasks that can change in the future, but it is very unlikely they will change in the next few years, such as historic sports statistics or other stable information, e.g., \emph{Which park is closest to the Rockefeller Archeological Museum and has a playground for children?}, $15.4\%$ of tasks, and (c) tasks that are unlikely to change in the next year, but can change in a longer period of time, e.g.,  business details as shown in Fig.~\ref{fig:trajectory}, accounting for $41.1\%$ of the tasks.

Tab.~\ref{tab:time_dependencys} in \S\ref{appendix_analysis} presents accuracy for the different splits with GPT4-T on the \assistbench{} test set. For all splits, trends are similar to those in Tab.~\ref{tab:test_results}, i.e.,  \cbones{} performs best and \seeactexplorer{} outperforms \seeact{}.
Between splits, accuracy on the \emph{static} and \emph{stable} splits is lower than $21$ points for all models, suggesting that these tasks can be a helpful evaluation resource in the next few years.

\section{Related Work}
\label{sec:rw}

\paragraph{Web agent benchmarks.}

In addition to the works described in \S\ref{sec:background}, other benchmarks explore web browsing scenarios that require dialogue \cite{lù2024weblinx}, following instructions from annotation tasks \cite{xu2024turkingbench}, and multi-modal understanding of web pages (e.g., OCR)  \cite{liu2024visualwebbench}.
Furthermore, recent datasets explore computer interaction beyond web usage, such as operating system management \cite{xie2024osworld, liu2024agentbench}, mobile applications \cite{rawles2023androidinthewild, toyama2021androidenv, rawles2024androidworld}, and work environments \cite{drouin2024workarena}. 

\paragraph{Web agents.}
Our focus is on web agents that are based on prompted foundation models due to their recent success \cite{yao2023react, zheng2023synapse, he2024webvoyager,zheng2024seeact, koh2024visualwebarena}. There has also been a growing interest in training specialized models for web interaction \cite{gur2018learning, nakano2022webgpt, pmlr-v162-humphreys22a, shaw2023from, gur-etal-2023-understanding, deng2023mindweb, furuta2024multimodal, gur2024a}. 
Other than web interaction, recent work explored LM agents that interact with predefined tools \cite{schick2023toolformer, Tang2023ToolAlpacaGT, qin2023tool, hao2023toolkengpt}, or perform other specific tasks, e.g., generate code \cite{yang2024sweagent}, play games \cite{Zhu2023GhostIT}, and operate embodied robots \cite{huang2022language}. 
\citet{liu2024agentbench} and \citet{kim2024biggen} evaluate LM agents on a wide range of tasks. For a survey on autonomous LM-based agents, see \citet{Wang_2024}.

\section{Conclusion}
We introduce \assistbench{}, a new challenging benchmark consisting of \numtasks{} tasks that tests the ability of web agents to assist with realistic and time-consuming tasks. 
\assistbench{} contains diverse tasks covering various scenarios and domains over more than \numuniquepages{} pages and \numuniquedomains{} websites.
In addition, we introduce \seeactexplorer{}, an improved web agent aimed to solve tasks in \assistbench{}. We compare \seeactexplorer{} with current web agents as well as strong closed-book and retrieval-augmented models and find that \assistbench{} is challenging for all models. We find that no model reaches an accuracy of more than $26$ points.
Closed-book models have currently the best accuracy, but they tend to hallucinate facts and suffer from low precision.
Our model, \seeactexplorer{}, significantly improves over current web agents and an ensemble that combines \seeactexplorer{} with a closed-book model reaches the best overall performance. Finally, we present a thorough analysis showing that web navigation remains a significant challenge. We hope \assistbench{} can serve as a useful benchmark to evaluate future web agents.

\section{Limitations}

\paragraph{Coverage of tasks.}
In \assistbench{}, we introduce a novel dataset consisting of useful and challenging web tasks. However, there were some realistic tasks that we were not able to include in the benchmark. 
When collecting data (\S\ref{dataset_criteria}) we received time-dependent tasks that cannot be automatically verifiable, such as: \emph{"Are there any tickets available for Billy Joel concerts this year that cost less than $\$100$ and aren't in a restricted viewing section?"}. 
Future work can expand \assistbench{} to time-dependent tasks, for example by extending specialized evaluation models \cite{kim2024prometheus, kim2024prometheus2} to a multi-modal web setting, which will allow automatic evaluation without reference answers.
In addition, tasks in \assistbench{} are limited to the participants we were able to contact.
While we were able to expand our benchmark to different domains, inspired by recent work in summarization \cite{hendrycks2021measuring} and question-answering \cite{malaviya-etal-2024-expertqa, malaviya2024dolomites}, future work can further scale our data collection method.

\paragraph{Benchmark size.}
\assistbench{} is smaller than some recent web agent benchmarks \cite{deng2023mindweb, zhou2024webarena}. Nevertheless, recent years saw a surge in popularity in small, high-quality benchmarks.
For example, \textsc{DrawBench} \cite{NEURIPS2022_ec795aea}, \textsc{HumanEval} \cite{chen2021evaluating} and \textsc{Bamboogle} \cite{press-etal-2023-measuring} are significant benchmarks with $200$, $164$, and $125$ examples respectively. Empirically, it has been shown that large datasets can be reduced to $100$ examples without compromising their effectiveness \cite{polo2024tinybenchmarks}. Contrary, a smaller benchmark also has some advantages: it results in a smaller environmental footprint \cite{10.1145/3381831} and has lower costs, potentially enabling more research groups to evaluate on our benchmark. This is especially important in an agent setup which is particularly expensive due to the high number of model calls. We discuss cost in more detail in \S\ref{appendix_models}.

\paragraph{Open-source web agents.} A limitation in the presented results is that our models are all based on GPT4-Turbo and Claude-3.5-Sonnet. While this limits reproducibility, it is in-line with recent works showing multi-modal web agents based on GPT-4 achieve state-of-the-art results \cite{zheng2024seeact, he2024webvoyager} with a large gap from open-source models \cite{liu2024visualwebbench}, and hence provides a good upper bound for performance on \assistbench{}.
We are currently experimenting with open-source models and are hopeful that, as open-source multi-modal models continue to improve in the near future \cite{liu2023visual}, \assistbench{} will be a useful evaluation resource. For example, future work can explore methods to create training data to improve open web navigation, a common failure of current models (\S\ref{sec:error_analysis}).

\section{Ethical Implications and Broader Impact}
\label{sec:ethical_implications}

Autonomous web agents have great potential to assist humans with useful tasks, such as those presented in \assistbench{}. However, they also pose potential risks. Agents capable of executing a wide range of tasks can impact the job market or help malicious users, e.g., to automatically spread fake news on the web. In addition, even if web agents become powerful to solve \assistbench{}, we will need to ensure their safe deployment to users. For example, they should not be able to share unintended personal details or make undesired transactions. Environmental considerations should also be taken into account, as web agents can incur high costs due to the large amount of model calls (see \S\ref{appendix_models} for a discussion on model costs).

\section*{Acknowledgments}
We would like to thank our colleagues at TAU NLP and The Allen Institute for AI for their insightful comments. Specifically, we would like to thank Maor Ivgi, Eden Avnat, Avia Efrat, and Tomer Wolfson for their help. We extend special thanks to Yu Su and Mor Geva for their insightful feedback. In addition, we thank SerpAPI for their support by granting us an academic discount. This research was partially supported by the Yandex Initiative for Machine Learning and the European Research Council (ERC) under the European Union Horizon 2020 research and innovation programme (grant ERC DELPHI 802800). This work was completed in partial fulfillment of the Ph.D. of the first author.

\bibliography{custom}

\appendix

\section{Appendix}
\label{sec:appendix}
We now provide additional examples technical details.

\subsection{Detailed Comparison to Recent Benchmarks}
\label{appendix_comparison}
Tab.~\ref{tab:benchmarks_examples} presents an example task for \assistbench{} and recent prominent web agent benchmarks.

\begin{table*}[t]
  \centering
  \footnotesize
  \begin{tabular}{ll} \toprule
    Dataset &  Example \\ \midrule

     \wob{} \cite{liu2018reinforcement} & Find the email by Ilka and forward it to Krista. \\ \midrule

    \webshop{} \cite{yao2022webshop} &  I would like a anti perspirant \\  
    & (Start from Shopping site) \\ \midrule
    
    \mindtoweb{} \cite{deng2023mindweb} & Show me a list of comedy movies, sorted by user ratings.   \\ 
        & (Start from IMDB) \\ \midrule

    \webarena{} \cite{zhou2024webarena} & What is the top-1 best-selling product in 2022?  \\ 
    & (Start from Shopping admin) \\ \midrule

    \webvoyager{} \cite{he2024webvoyager} & Search for an Xbox controller with green color and rated above 4 stars  \\
    & (Start from Amazon) \\ \midrule

    \mmina{} \cite{zhang2024mmina} & When did programming language that has the largest variance in salary first appear? \\ & Answer using the information from the Wikipedia site in the second tab. \\ \midrule

    \gaia{} \cite{mialon2024gaia}  & In the year 2022, and before December, what does "R" stand for in the three core \\ & policies of the type of content that was violated in the public logs on the \\ 
    &  Legume Wikipedia page? \\ \midrule

        \assistbench{} (ours) & What’s the highest price a high-rise apartment was sold for in Mission Bay, San \\
        & Francisco, in 2021? \\

    \bottomrule

  \end{tabular}
  \caption{An example task from \assistbench{} and other prominent web agent benchmarks.} 
     \label{tab:benchmarks_examples}

\end{table*}

\subsection{Seed Data Collection}
\label{appendix_seed}

\begin{figure}[t]
  \includegraphics[width=1.0\columnwidth]
{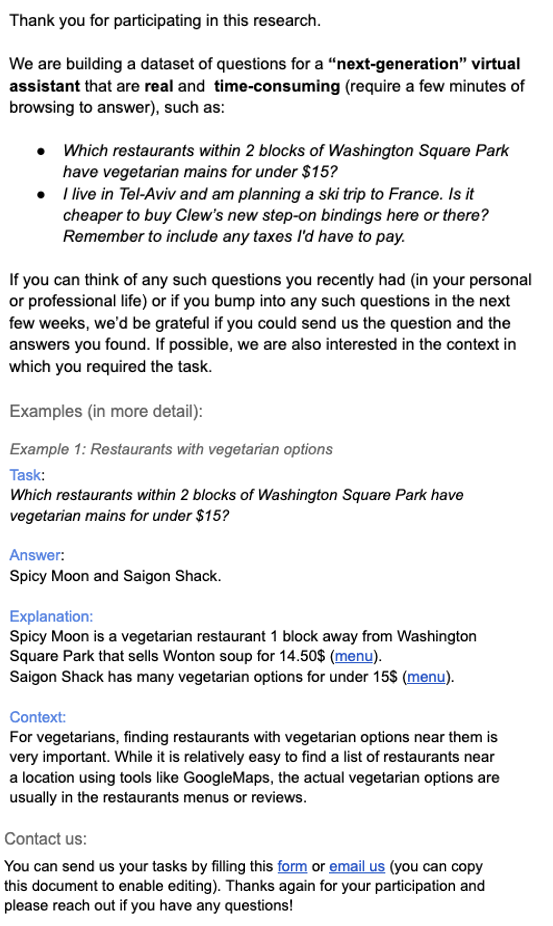}
  \vspace*{-0.7cm}
  \caption{The instructions shown to participants in our data collection survey.}
  \label{fig:survey_instructions}
  \vspace{-10pt}
\end{figure}

Fig.~\ref{fig:survey_instructions} presents the instructions shown to participants in the seed data collection survey. We shared the survey with $20$ participants and received $128$ tasks. $56$ tasks were filtered for not adhering to the benchmark dataset criteria (\S\ref{dataset_criteria}). Filtered tasks include time-dependent tasks, e.g., \emph{``Where can I get the cheapest ticket for the musical lion king for tonight in the main stage area NYC?''}, and tasks with many answers, e.g., \emph{``I am going on a trip to Barcelona, which venues in the city have Gaudi statues and are free to visit?''}. 

When validating the tasks, we made small tweaks to ensure the task can be automatically verified. For example, we added a location to the received task: \emph{``Where is the nearest outlet that has as much of these stores as possible [Columbia, Lululemon, Ted Baker, and Michael Kors]''} to \emph{``Which outlets in Ontario have all of the following stores: Columbia, Lululemon, Ted Baker, and Michael Kors?''}. At the end of the process, we are left with $72$ tasks from \numgeneralparticipants{} participants, an average of $4$ tasks per participant. As shown in Fig.~\ref{fig:datacollection}, there is a variance in tasks between participants, as different participants have different interests. We found direct communication with the participants to be helpful to explain the task guidelines and to validate received tasks.

\subsection{Expanding the Seed Set with Crowd-workers}
\label{appendix_expanding_set}
\begin{figure*}[t]\setlength{\belowcaptionskip}{-8pt}
  \centering
  \includegraphics[width=\textwidth]{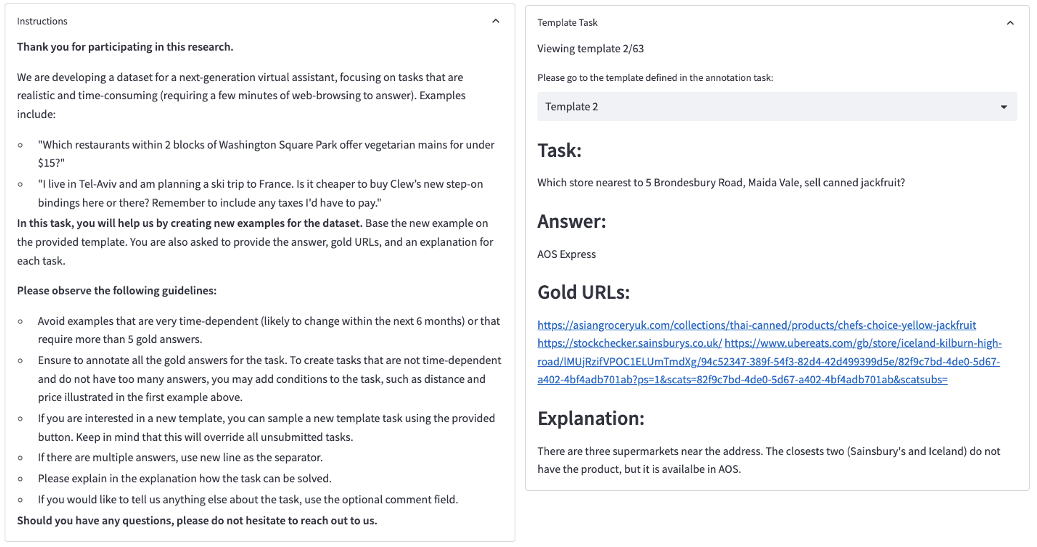}
  \vspace*{-0.7cm}
  \caption{Snapshots from our data collection server we used to expand the seed set. (left) The instructions shown to the annotators. (center) An example template from the seed set.}
  \label{fig:duplicate_seed}
\end{figure*}

To expand the seed set, we asked crowd-workers to use tasks from the set as templates for new tasks. Fig.~\ref{fig:duplicate_seed} presents our data collection server\footnote{We use Streamlit, \url{https://streamlit.io/,} to build the server}, including our instructions. Due to the complexity of the task, we first ran a qualification task with five experienced annotators and continue with the two annotators that performed best. The average time to complete a task was between $20$ to $30$ minutes and we pay workers $\$7$ per example. To verify data quality, we published batches of between $10$ to $15$ examples and were in direct contact with the workers.

\subsection{Collecting Examples from Domain-experts}
\label{appendix_expert_collection}
\begin{figure*}[t]\setlength{\belowcaptionskip}{-8pt}
  \centering
  \includegraphics[width=\textwidth]{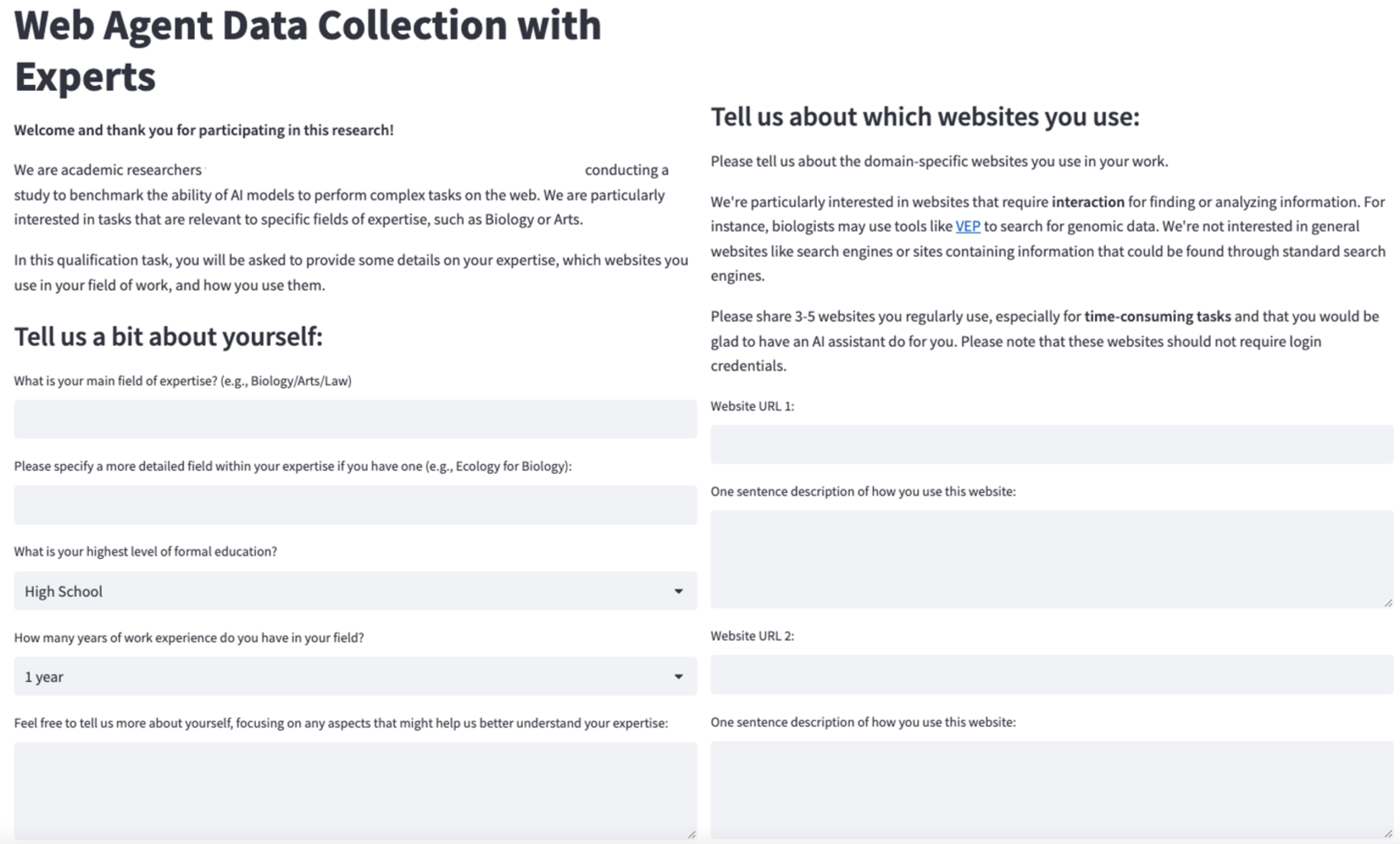}
  \vspace*{-0.7cm}
  \caption{Snapshots from our data collection server we used to for the expert qualification task. (left) Participants share background details about their professional experience. (right) Participants share information about professional websites and how they use them.}
  \label{fig:expert_qualification}
\end{figure*}

\begin{figure*}[t]\setlength{\belowcaptionskip}{-8pt}
  \centering
  \includegraphics[width=\textwidth]{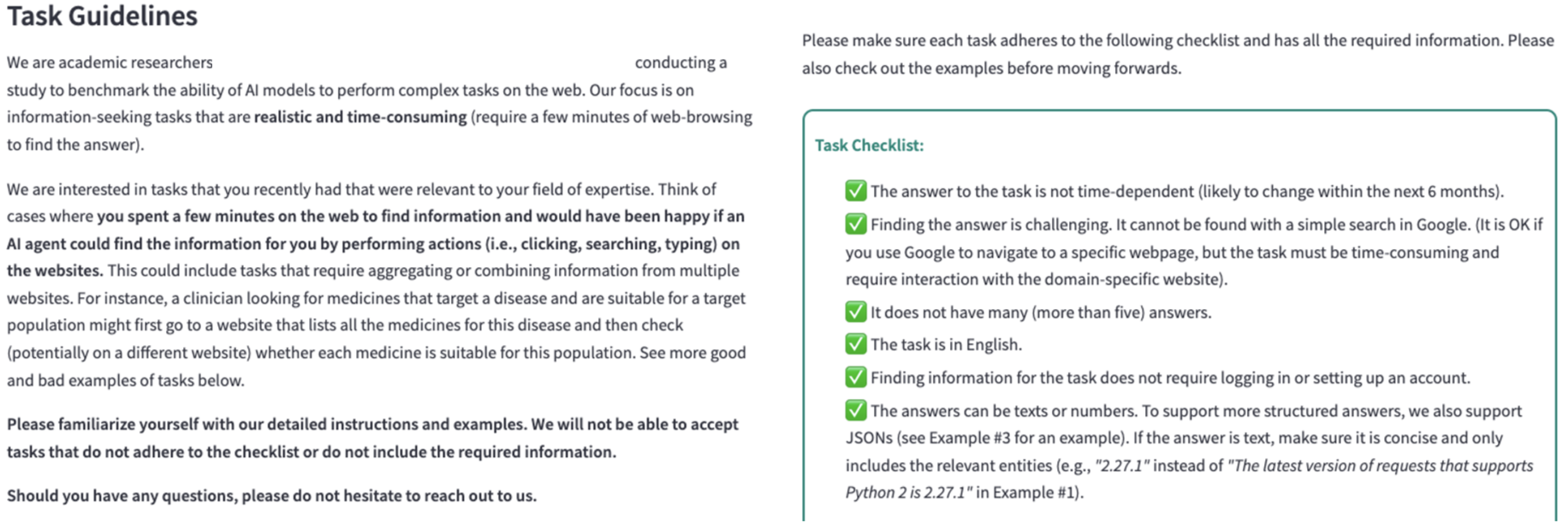}
  \vspace*{-0.7cm}
  \caption{Snapshots from our data collection server we used to for the expert task. (left) Task guidelines (right) Checklist submitted for submitted task. We also show participants an instruction video, in addition to examples, detailed explanation about the annotation format, and feedback from the qualification task, which we leave out for brevity.}
  \label{fig:expert_annotation}
\end{figure*}

We recruited 35 participants from the Prolific crowdsourcing platform. Participants were required to be fluent in English and came from various countries across Europe, the Americas and Africa. They also needed to have at least 99\% approval rate on Prolific and at least 50 prior approved submissions. Participants were explicitly told about the goals of our study and how their data would be used. We paid participants \$20/hour in both the qualification and the main study. We allocated 5 minutes for the qualification study and 15 minutes for the main study where they provided a task.
Fig.~\ref{fig:expert_qualification} and Fig.~\ref{fig:expert_annotation} present snapshots from our qualification and annotation task, respectively.

\subsection{\assistbench{} Development Set}
\label{appendix_dev_set} 

\begin{table}[t]
\centering
\footnotesize
  \begin{tabular}{lc}
    \toprule
    Model  & Cost (\$) \\ \midrule 
    \cbinst{} & 0.006\\
    \cbones{} &  0.007\\
    \ralminst{} & 0.049\\
    \ralmones{} & 0.069\\
    \seeact{} & 2.094 \\  \midrule
    \textsc{\seeactexplorer{}} (ours) & 2.472  \\
  \bottomrule
\end{tabular}

\caption{Average cost to evaluate an example on the \assistbench{} test set.}
  \label{tab:costs}
\end{table}

Tab.~\ref{tab:full_dev_set} shows example tasks from the \assistbench{} development set.

\subsection{Models}
\label{appendix_models}

\paragraph{GPT4.}
We use GPT-4-turbo-2024-04-09 as the backbone for all our models. 
\paragraph{Costs.}

Tab.~\ref{tab:costs} presents the average cost to evaluate an example on the \assistbench{} test set. Multi-modal web agents are significantly more expensive than closed-book and retrieval-augmented LMs.

\paragraph{Retrieval-augmented models.}

Our retrieval-augmented models are limited to ten searches (see \S\ref{appendix_analysis} for an analysis on accuracy as a function of number of steps). Our retriever is \google{}\footnote{We query \google{} via SerpAPI: \url{https://serpapi.com}}.
For \fanoutqa{}, we prepend \emph{'en.wikipedia.com'} to every retrieval query, following recent work \cite{yoran2023answering, yoran2024making} and prepend the retrieved context to the task in all settings \cite{trivedi2022interleaving, yoran2023answering}.

\paragraph{\seeact{}.}

We used SeeAct from the original paper \cite{zheng2024seeact} with two minor changes: 1. At the beginning of each step we change the google settings of region and language to be `US', 2. We increase the maximum number of steps to 30, similarly to \seeactexplorer{}. 
Since \seeact{} does not generate answers, we manually extract the answer from the \emph{"thoughts"} of the model.

\paragraph{\seeactexplorer{}.}

In addition to the planning and memory components described in \S\ref{sec:model}, we add new actions to support open web naviation. These include: \emph{SCROLL} (scroll down to 75\% of the current view or up to the top of the page), \emph{GOTO} (go to a specific URL), \emph{SEARCH} (execute a search in Google), and \emph{GOBACK} (go back to previous page). 

Our web agents operate by taking screenshots of the web page and passing them to the multi-modal model. Since they operate in the open-web they are subject to technical failures. For example, occasionally the agent fails to take a screenshot which causes the run to crash. We use the same Playwright library as the original \seeact{}.\footnote{Playwright Library: \url{https://playwright.dev/docs/api/class-playwright}}

\subsection{\fanoutqa{}}
\label{sec:appendix_fanout}
We use tasks from \fanoutqa{} \cite{zhu2024fanoutqa} as an additional development set that requires reasoning on a single site, Wikipedia.
The \fanoutqa{} development set includes $310$ tasks, with the vast majority ($288/310$) having dictionary answers which are supported in our setting.

We make several design choices to use \fanoutqa{} in our setting. First, we remove tasks with more than $5$ answers and are left with $210$ unique tasks. Then, we convert the answer format to our evaluation framework (\S\ref{subsec:eval}) by changing the answers from a single dictionary where the keys are entities and the values are the matching values for each entity (e.g., \emph{`\{"Albania": 67, "Croatia":  40...\}'} to a list of dictionaries where each item has the same keys (e.g. \emph{`[\{"Country": "Albania", "Human Development Index": 67\}, \{"Country": "Croatia", "Human Development Index": 40\},...]'}), for the question \emph{`What is the Human Development Index ranking of NATO member states that joined NATO after 2005?'}

Then, to better differntiate between different web agents, we remove cases where the answer is already memorized by the parameters of our LM by running our \cbones{} model and filtering tasks where accuracy is larger than $0.5$. This step causes the relatively low support and we are left with only $60$ tasks. Finally, we manually go over these tasks and remove tasks that are ambiguous (e.g., ask about a country's GDP without mentioning if per capita or not), outdated, or when the predicted answer for \cbones{} is a paraphrase of the reference answer. We also make small refinements if tasks can be easily disambiguated.
At the end of the process, we are left with the $31$ tasks we use in \S\ref{sec:experiments}.

\begin{figure*}[t]\setlength{\belowcaptionskip}{-8pt}
  \centering
  \includegraphics[width=\textwidth]{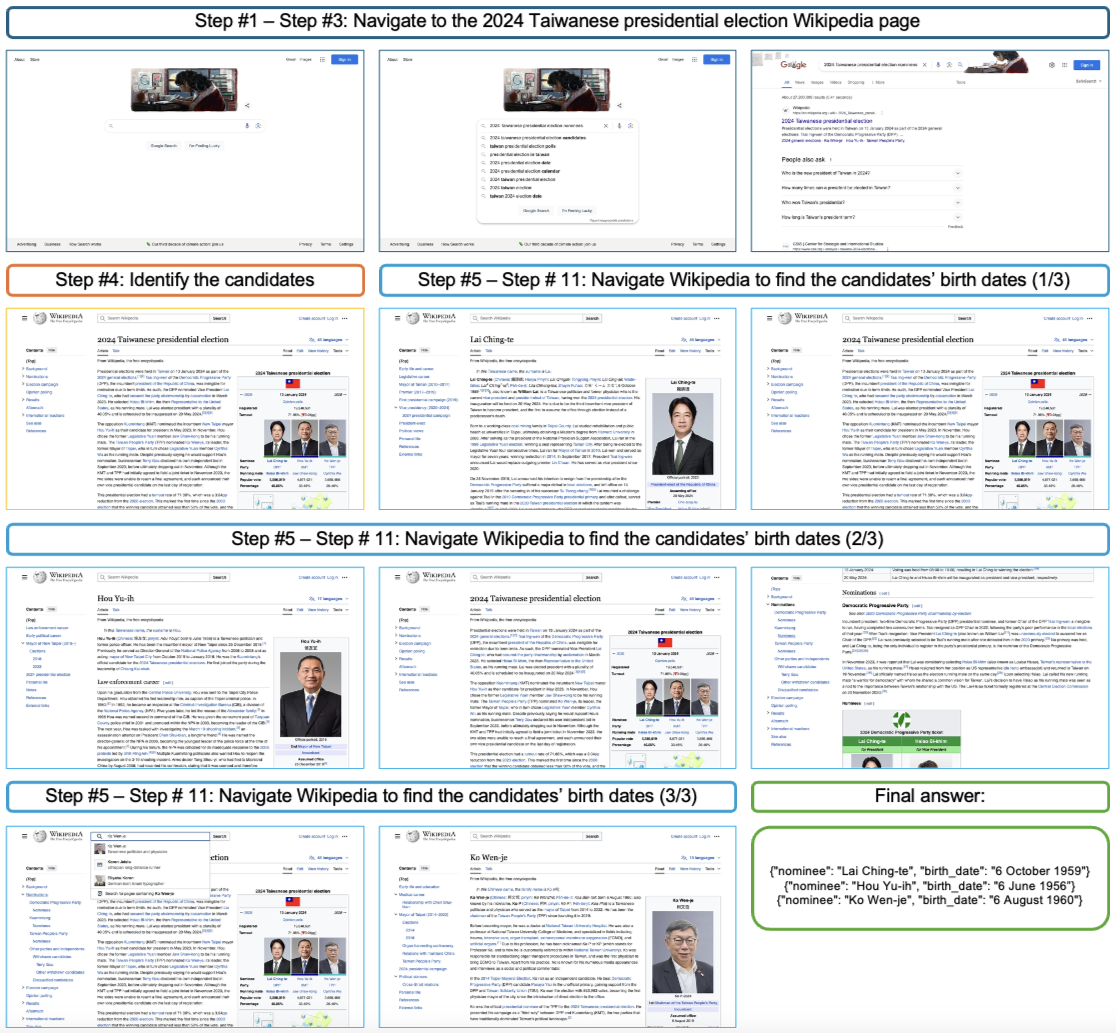}
  \vspace*{-0.7cm}
  \caption{An example for a success for \seeactexplorer{} on \fanoutqa{}. The model first navigates to the relevant Wikipedia page, and then fans out to find the relevant fact about each candidate, before generating the final answer.}
  \label{fig:fanout_success}
\end{figure*}

Because all models score very low on \assistbench{} and due to the high cost in running web agents (\S\ref{appendix_models}) we found this set useful in running small-scale experiments. Fig.~\ref{fig:fanout_success} presents an example where \seeactexplorer{} succeeds in generating the gold answer.

\subsection{Results}
\label{appendix_results}

Tab.~\ref{tab:test_results_claude} presents results with Claude-3.5-Sonnet as the engine for our models.
Tab.~\ref{tab:dev_results} Tab.~\ref{tab:difficulty_levels} and present results with GPT4-T on our development set and on the different difficulty levels on the test set.   
Tab.~\ref{tab:full_results_splits} presents the answer rate, precision, and exact match for the different types of collected data. All results reported are for a single run, due to the high costs in running experiments.


\begin{table*}[t]
\centering
\footnotesize
  \begin{tabular}{llcccc}
    \toprule
    Type & Model  & Acc. & Ans. $\%$ & Prec. & EM
    \\ \midrule 
    Closed-book LMs& \cbinst{} & 17.7 & 69.1 & 25.6 & 6.1 \\
    & \cbones{} & 21.9 & 76.2 & 28.8 & 6.6 \\ \midrule
    Retrieval-augmented LMs & \ralminst{} & 11.5 & 43.1 & 26.7 & 5.0 \\
    &  \ralmones{} & 11.0 & 42.5 & 25.9 & 3.3   \\ \midrule
   Web agents & \seeact{} & 2.2 & 13.8 & 15.8 & 1.7 \\ 
    & \seeactexplorer{} (ours) & 12.9 & 34.3 & \textbf{37.7} & 8.8\\ 
  \midrule     
  Ensembles & \ralminstfallback{} & 22.5 & 79.6 & 28.3 & 8.3  \\ 
      & \ralmonesfallback{} & 21.6 & 82.3 & 26.3 & 6.6 \\ 
     & \seeactfallback{} & 22.3 & 76.2 & 29.3 & 7.7  \\ 
     & \spafallback{} (ours) &  \textbf{26.4} & \textbf{81.8} & 32.2 & \textbf{13.8} \\ 
  \bottomrule
\end{tabular}
\caption{Results on the \assistbench{} test set with Claude-3.5-Sonnet.}
  \label{tab:test_results_claude}
\end{table*}

\begin{table}[t]
\centering
\footnotesize
  \begin{tabular}{lcccc}
    \toprule
    Model  & Acc. & Ans. $\%$ & Prec. & EM 
    \\ \midrule 
    \cbinst{} & 18.1 & 42.4 & 42.6 & 6.1 \\
    \cbones{} & 23.7 & 90.9 & 26.1 & 9.1 \\ \midrule
    \ralminst{} & 15.5 & 54.5 & 28.4 & 9.1 \\
    \ralmones{} & 13.6 & 42.4 & 32.1 & 6.1 \\  \midrule
    \seeact{} & 0.0 & 3.0 & 0.0 & 0.0 \\
    \textsc{\seeactexplorer{}} (ours) & 12.7 & 24.2 & \textbf{52.6} & 9.1 \\
    \midrule
    \ralminst{}$\rightarrow$\textsc{CB} & 28.1 & \textbf{93.9} & 29.9 & \textbf{12.1} \\ 
     \ralmones{}$\rightarrow$\textsc{CB} & 25.1 & \textbf{93.9} & 26.7 & 9.1 \\ 
    \seeact{}$\rightarrow$\textsc{CB} & 23.7 & \textbf{93.9} & 25.3 & 9.1 \\ \seeactexplorer{}$\rightarrow$\textsc{CB} (ours) & \textbf{29.3} & 90.9 & 32.3 & \textbf{12.1} \\ 
  \bottomrule
\end{tabular}

\caption{Results on the \assistbench{} development set with GPT4-T.}
  \label{tab:dev_results}
\end{table}

\begin{table}[t]
\centering
\footnotesize
  \begin{tabular}{lccc}
    \toprule
    Model  & Easy & Medium & Hard \\ \midrule 
    \cbinst{} & 51.2 & 40.2 & 2.3  \\
    \cbones{} & \textbf{82.1} & \textbf{49.7}  & 4.2  \\ \midrule 
    \ralminst{} & 50.2&  17.0 & 6.2\\
    \ralmones{} & 80.0&  10.5  & 5.5 \\ \midrule
    \seeact{} &28.9 & 2.5 & 2.9 \\ 
    \textsc{\seeactexplorer{}} (ours) &  29.5& 12.3 & 9.1 \\
    \midrule
    \ralminst{}$\rightarrow$\textsc{CB} & 57.8& 34.7 & 8.0 \\ 
     \ralmones{}$\rightarrow$\textsc{CB} & 81.3& 35.0 & 7.3 \\ 
    \seeact{}$\rightarrow$\textsc{CB} & 82.0 & 47.7 & 7.1 \\ \seeactexplorer{}$\rightarrow$\textsc{CB} (ours) & 80.7 & 42.7 &  \textbf{12.4} \\ \midrule
    \# of tasks  & 9 & 56 & 116 \\ \bottomrule
\end{tabular}

\caption{Accuracy on the \assistbench{} test set for different difficulty levels. The majority of the questions are hard for all models, and \seeactexplorer{} significantly outperforms \seeact{} on medium and hard tasks.}
  \label{tab:difficulty_levels}
\end{table}

\begin{table*}[t]
\centering
\footnotesize
  \begin{tabular}{l|ccc|ccc|ccc}
     & & Answer $\%$ & & & Precision & & & EM & \\
    Model & (seed) & (expanded) & (experts)  & (seed) & (expanded) & (experts)  & (seed) & (expanded) & (experts)\\ \toprule
    
    \cbinst{} & 55.7 & 53.9 & 40.5 & 28.3 & 36.3 & 28.0 & 7.1 & \textbf{6.9} & 2.4  \\
    \cbones{} & \textbf{97.1} & \textbf{92.2} & \textbf{71.4} & 30.5 & 20.2 & 27.2 & \textbf{14.3} & 4.9 &  \textbf{7.1}\\
    \ralminst{}{} & 57.1 & 67.6 & 42.9 & 23.5 & 18.7 & 22.7 & 7.1 & 5.9 & 4.8 \\
    \ralmones{} & 50.0 & 46.1 & 45.2 & 15.9 & 29.0 & 25.2 & 1.4 & \textbf{6.9} & 2.4 \\
    \seeact{} & 10.0 & 11.8 & 23.8 & 20.0 & 16.5 & \textbf{40.0} & 1.4 & 1.0 & 4.8 \\  \midrule
    \seeactexplorer{} (ours) & 32.9 & 30.4 & 45.2 & \textbf{39.4} & \textbf{38.7} & 16.9 & 7.1 & 5.9 & 4.8 \\
    
  \bottomrule
\end{tabular}
\caption{All metrics for our model on the different types of collected data. Accuracy is presented in Tab.~\ref{tab:expert_split}.}
  \label{tab:full_results_splits}
\end{table*}

\renewcommand{\arraystretch}{1.15}
\begin{table*}[t]
    \centering
    \footnotesize
    \rowcolors{2}{gray!15}{white}
    \begin{tabular}{p{0.75\linewidth} p{0.2\linewidth}} \toprule
    \rowcolor{gray!30}
        Question & Answer \\ \midrule
        Which gyms near Tompkims Square Park (<200m) have fitness classes before 7am? & CrossFit East River, Avea Pilates East Village\\
        Based on recent years (2020-2023), how likely am I to hit a rainy day (at least 0.5mm of precipitation) if I travel to Seattle during the first week of September? (provide the answer in percentage) & 14.20\% \\
        What's the lowest price a Single Family house was sold in Queen Anne in January 2023? & 1010000\$\\
        Which restaurants (not takeaway only) within 1 blocks of Washington Square Park have vegan mains for under \$15? & Shanghai villa\\
        Which paintball places in Cologne, Germany are within a 10 minute walk from a karting track? & Adrenalinpark Köln\\
        What is the highest rated (according to IMDB) Daniel Craig movie that is less than 150 minutes and is available on Netflix (US)? & Glass Onion: A Knives Out Mystery\\
        How much does it cost to send an envelope with 1-week delivery from Rio de Janeiro to NYC with DHL, USPS, or Fedex? (Format your answers as a list of jsons with the keys "sender" and "price (usd)" for each option you find) & \{"sender": "DHL", "price (usd)": "55-70"\} \\
        Which Vanguard ETF had the highest pct increase between 2013 and 2023: VGT, MGK, or VONG? & VGtT\\
        What's the smallest house (based on square footage) that has at least 2 beds and 2 baths and was sold in Prince Edward Island between June 1, 2022 and May 15 2024 according to Zillow? & 1148 sqft\\
        What is the cheapest option to mail a DVD to Colombia from Hartford, Connecticut using FedEx, DHL, or USPS? (The answer should be a json object with the keys "sender" and "price (usd)") & \{"sender": "USPS", "price (usd)": "41.75"\} \\
        Which bar is closest to Mummers Museum in Philadelphia and is wheel chair accessible? & For Pete's Sake\\
        What is the highest rated (according to IMDB) Isabelle Adjani feature film that is less than 2 hours and is available on Vudu (now called Fandango at Home) to buy or rent? & Nosferatu the Vampyre\\
        Which members of Fubo's Management Team joined the company during the same year Fubo's IPO happened? & Gina DiGioia\\
        How much will I save by getting annual passes for a group of 4 adults and 1 student for the Philadelphia Museum of Art, compared to buying daily tickets, if we visit 5 times in a year (all non-consecutive days)? & \$395\\
        What is the closest eatery to Harkness Memorial State Park that is still open at 11pm on Wednesdays? & McDonald's \\
        Where can I take martial arts classes within a five-minute walk from the New York Stock Exchange after work (7-9 pm)? & Renzo Gracie Jiu-Jitsu Wall Street \\
        What popular hiking trails to waterfalls in Yosemite National Park (more than 1,000 reviews on TripAdvisor) have been recommended to be fully accessible to wheelchairs by at least three different people and are highly rated (4.5/5 or more on average)? & Yosemite Falls, Bridalveil Fall \\
        Which supermarkets within 2 blocks of Lincoln Park in Chicago have ready-to-eat salad for under \$15? & Potash Markets - Clark Street \\
        What is the worst rated series (according to Rotten Tomatoes) with more than 1 season that Ted Danson has starred in and is available on Amazon Prime Video (US)? & CSI: Cyber\\
        Which Fidelity international emerging markets equity mutual fund with \$0 transaction fees had the lowest percentage increase between May 2019 to May 2024? & Fidelity® Emerging Markets Index Fund (FPADX) \\
        \bottomrule
    \end{tabular}
  \caption{Examples from the \assistbench{} development set. In addition to the answer, we also collect the URLs where the answer can be verified and an explanation on how the task can be solved, which we do not include here for brevity. If the answer consists of more than one dictionary, only one is shown.} 
     \label{tab:full_dev_set}
\end{table*}

\subsection{Examples and Analysis}
\label{appendix_analysis}

 \begin{figure}[t]
  \includegraphics[width=1.0\columnwidth]{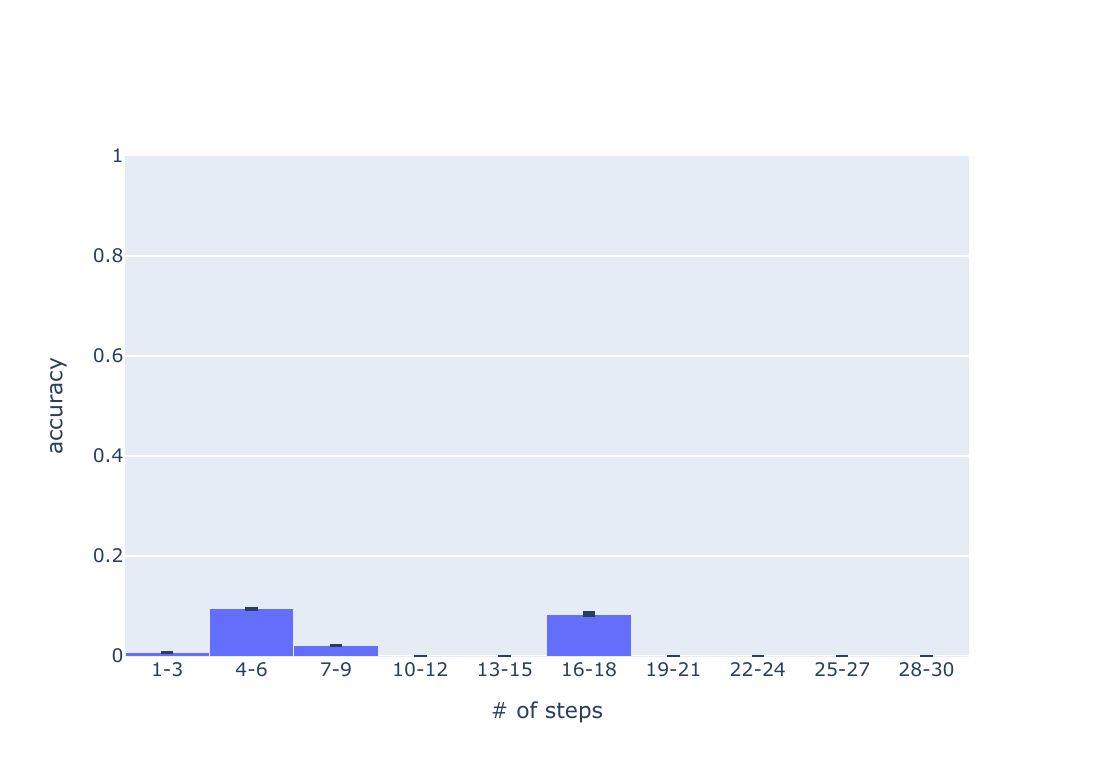}  
  \caption{Accuracy on the \assistbench{}{} test set for \seeact{} as a function of the number of execution steps. Error bars indicate standard error of the mean. Web agents struggle when trajectories are very long or very short.}
  ~\label{fig:sa_num_calls}
  \vspace{-10pt}
\end{figure}

\begin{figure*}
\begin{subfigure}{.5\textwidth}
  \centering
  \includegraphics[width=1\linewidth]{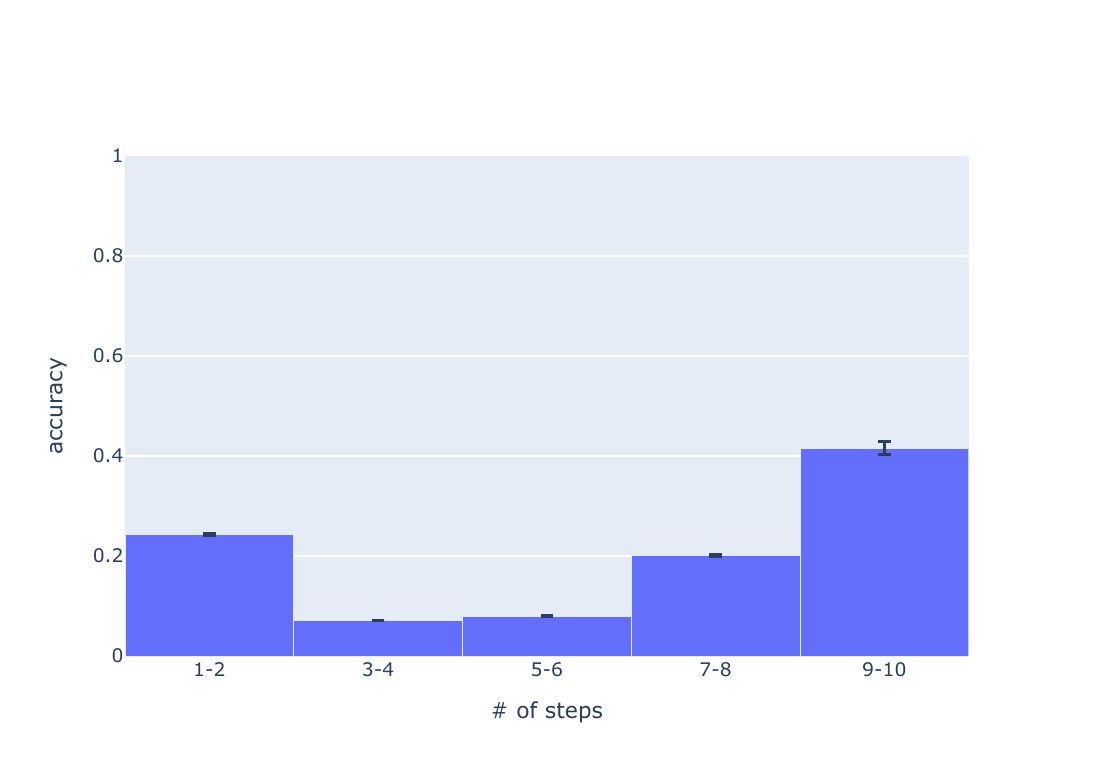}  
\end{subfigure}
\begin{subfigure}{.5\textwidth}
  \centering
  \includegraphics[width=1\linewidth]{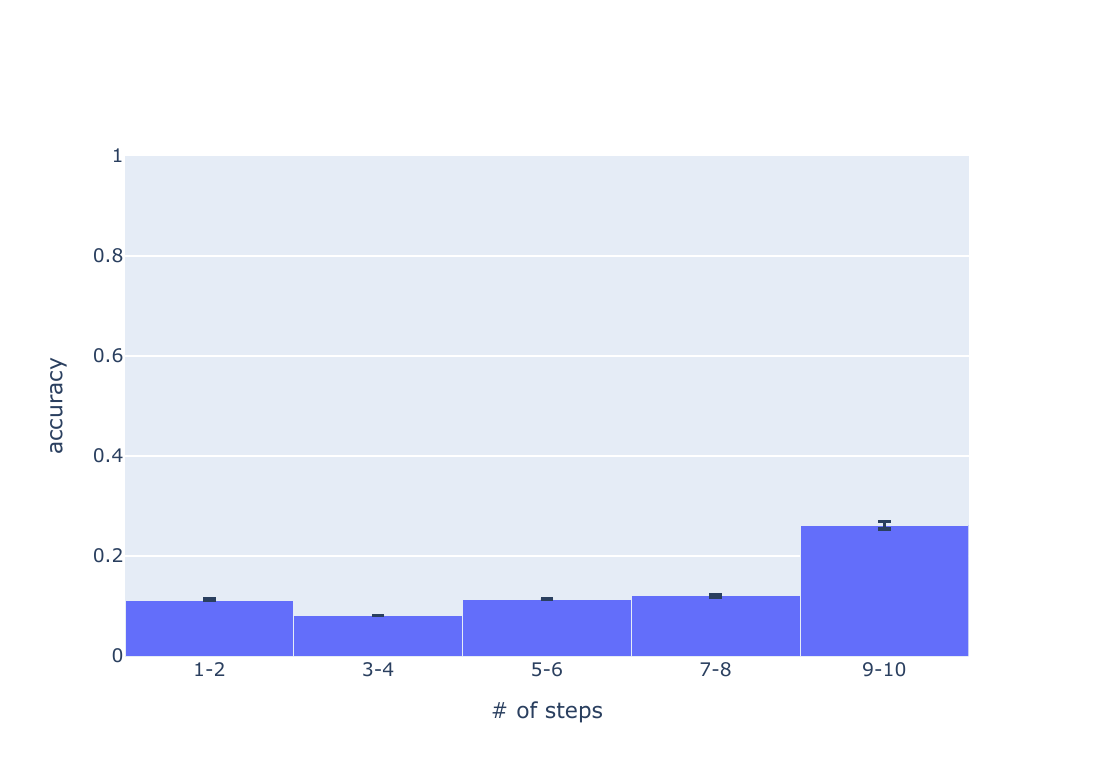}  

\end{subfigure}

\caption{Accuracy on the \assistbench{} test set for our \ralminst{} (left) and \ralmones{} (right) as a function of the number of execution steps. Error bars indicate standard error of the mean. Retrieval-augmented LMs struggle when trajectories are very long or very short, similarly to our web agents (Fig.~\ref{fig:seeact_num_calls} and Fig.~\ref{fig:sa_num_calls}).}
\label{fig:num_steps}
\end{figure*}

\begin{figure*}[t]
  \centering
  \includegraphics[width=\textwidth]{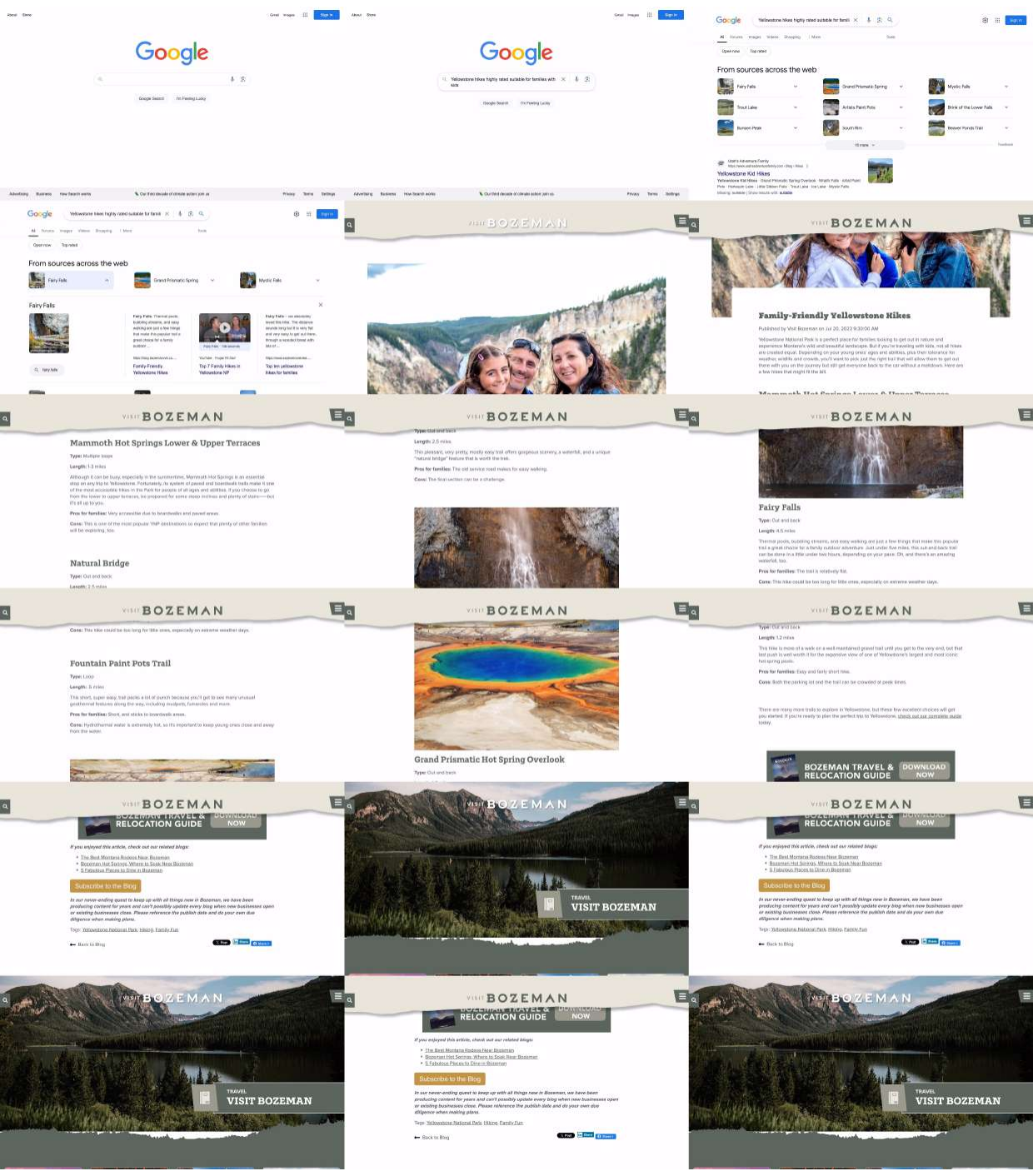}
  \vspace*{-0.7cm}
  \caption{An example of a navigation failure of \seeactexplorer{}. The model scrolls down and up the same page in a loop. Our analysis shows that long trajectories often lead to errors (\S\ref{sec:quantitative_analysis}).}
  \label{fig:navigation_failure}
\end{figure*}

\begin{figure*}
\colorbox{gray!10}{
\begin{minipage}{15cm}
\texttt{\textbf{\emph{No answer - proposes plan}} \\
\textbf{Task:} \emph{The dog genome was first mapped in 2004 and has been updated several times since. What is the link to the files that were most relevant in May 2020?}
\textbf{Model's answer:} \emph{Thought: To find the link to the dog genome files most relevant in May 2020, it would be ideal to check the archives of genomic databases or repositories such as NCBI, Ensemble or UCSC Genome Browser around that time. However, as of my last training data, I cannot browse the internet or access real-time data to provide a current link.\\
The final answer is:}}  \\ 
\texttt{\textbf{\emph{Wrong answer - hallucinated element}} \\
\textbf{Task:} \emph{How much will I save by getting annual passes for my family (2 adults, 1 kid age 5, 1 kid age 2) for the Seattle Children Museum, compared to buying daily tickets, if we visit 4 times in a year?} \\
\textbf{Model's answer:} \emph{Thought: First, I need to find the cost of daily tickets and annual passes at the Seattle Children's Museum for the family configuration provided. Then, I will calculate the total cost for both options if the family visits 4 times in a year and compare them to find the amount saved. \\
From the Seattle Children's Museum: \\
\textit{- Daily ticket prices are:} \\
  - Adults: \$12 each \textit{(hallucination)} \\
  - Children (1 year and older): \$12 each \textit{(hallucination)} \\
  - Children under 1 year: Free \\
\textit{- Annual pass prices are:} \\
  - Family (up to 2 adults and all children in the household): \$135 \textit{(hallucination)} \dots \\
}}
\texttt{\textbf{\emph{Wrong answer - outdated info}}} \\
\texttt{\textbf{Task:} \emph{What is the highest rated (according to IMDB) Daniel Craig movie that is less than 150 minutes and is available on Netflix (US)?}} \\
\texttt{\textbf{Model's answer:} \emph{Thought: To find the highest rated Daniel Craig movie on IMDB that is less than 150 minutes and available on Netflix (US), I need to: \\
1. Identify movies featuring Daniel Craig.\\
2. Check the runtime of these movies to ensure they are under 150 minutes.\\
3. Look up their ratings on IMDB.\\
4. Verify their availability on Netflix (US).\\ \dots}} \\
\texttt{ 4. **Netflix (US) Availability**: As of the last update, "Casino Royale," "Skyfall," and "Knives Out" have been available on Netflix at different times \textbf{(outdated, Casino Royale is not on Netflix anymore)}. Current availability can vary due to changes in Netflix’s catalogue.\\
Assuming all information is up-to-date, and these movies are still available on Netflix (US), the highest-rated Daniel Craig movie that satisfies all conditions is "Casino Royale" with an IMDB rating of 8.0. \\
The final answer is: Casino Royale}

\end{minipage}}
\caption{Example failures for our closed-book LMs.}
\label{fig:closed_book_failures}
\end{figure*}

\begin{figure*}
\colorbox{gray!10}{
\begin{minipage}{15cm}
\texttt{\textbf{\emph{Bad retrieval - map-related}} \\
\textbf{Task:} \emph{Which bar is closest to Mummers Museum in Philadelphia and is wheel chair accessible?} \\
\textbf{Search:} \emph{closest wheelchair accessible bar to Mummers Museum Philadelphia}  \\ 
\textbf{Result:} \emph{The Best 10 Bars near Mummers Museum in Philadelphia ...: Reviews on Bars near Mummers Museum - Lucky's Last Chance, Fountain Porter, The Dive Bar, Royal Tavern, New Wave Cafe, 2nd Street Brew House, Friendly ... \textbf{None of these are the closest to Mummers Museum}}} \\
\texttt{\textbf{\emph{Bad retrieval - partial information}} \\
\textbf{Task:} \emph{Based on recent years (2020-2023), how likely am I to hit a rainy day (at least 0.5mm of precipitation) if I travel to Seattle during the first week of September? (provide the answer in percentage)} \\
\textbf{Search:} \emph{average rainfall days in Seattle first week of September 2020-2023} \\
\textbf{Result:} 2023 Rainfall: Normal annual rainfall for Seattle is 39.34. Seattle rainfall by month in 2023  Seattle Weather Blog. ABOUT US. The Seattle Weather Blog provides weather ...} \\
\texttt{\textbf{\emph{Bad retrieval - retrieval error}} \\
\textbf{Task:} \emph{What's the lowest price a Single Family house was sold in Queen Anne in January 2023?} \\
\textbf{Search:} \emph{lowest price Single Family house sold Queen Anne January 2023} \\
\textbf{Result:} Queen Anne, Seattle Housing Market: House Prices \& Trends: In April 2024, Queen Anne home prices were up 11.7\% compared to last year, selling for a median price of \$1.0M. On average, homes in Queen Anne sell after 16 ...}

\end{minipage}}
\caption{Example failures for our retrieval-augmented LMs.}
\label{fig:ralm_failures}
\end{figure*}

\paragraph{Time-dependency of tasks.}
Tab.~\ref{tab:time_dependencys} presents results for the different time-dependency splits described in \S\ref{sec:time_dependency}.

\begin{table}[t]
\centering
\footnotesize
  \begin{tabular}{lccc}
    \toprule
    Model  & Accuracy & Accuracy & Accuracy \\
      & (static) & (stable) & (unlikely)
    \\ \midrule 
    \cbinst{}  & 15.4 & 13.9 & 18.8\\
    \cbones{} & \textbf{20.4} & 
    \textbf{18.1} & \textbf{25.9} \\
    \ralminst{} & 14.7 & 9.7 & 9.4\\
    \ralmones{}  & 11.0 & 13.0 & 9.5\\
    \seeact{} & 3.3 & 12.2 & 1.4 \\ 
    \seeactexplorer{} & 8.7 & 16.5 & 11.4 \\ \midrule
    Avg. & 12.3 & 13.9 & 12.7 \\
  \bottomrule
\end{tabular}
\caption{Accuracy on the \assistbench{} test set for different time-dependency splits with GPT4-T.}
  \label{tab:time_dependencys}
\end{table}

\paragraph{Accuracy as a function of the number of execution steps.}
Fig.~\ref{fig:sa_num_calls} and Fig.~\ref{fig:num_steps} presents the accuracy for \seeact{} and retrieval-augmented models as a function of the number of steps, respectively.

\paragraph{Closed book failures examples.}
 Fig.~\ref{fig:navigation_failure} shows a \emph{web navigation} error for \seeactexplorer{} where the model scrolls up and down the same page in a loop.
 Fig.~\ref{fig:closed_book_failures} and Fig.~\ref{fig:ralm_failures} show example for different failure cases for our closed-book and retrieval-augmented models, respectively.

\subsection{Prompts}
\label{appendix_prompts}

Fig.~\ref{fig:cb_prompt} shows the prompt for the closed book setup.
Fig.~\ref{fig:ralm_prompt} shows the prompt for the retrieval-augmented models.
The different prompts for \seeactexplorer{} can be found in Fig.~\ref{fig:spa_sys_prompt}, Fig.~\ref{fig:spa_first_turn_prompt}, and Fig.~\ref{fig:spa_second_turn_prompt}.

\begin{figure*}
\colorbox{gray!10}{
\begin{minipage}{15cm}
\texttt{\textbf{\emph{System prompt:}} You are a helpful agent that has to provide answers to tasks.
Don't ask the user any questions. \\
When done, end by generating a new line, followed by the final answer. \\
The format for the final answer should be The final answer is: {ANSWER} with ANSWER being the answer to the task. \\
Keep the answer as concise as possible. The answer should either be: a number, a string, a list of strings, or a list of jsons. \\
The answer should be parsed with the python method: json.loads(input\_str). If no answer is found, generate an empty string. \\
At each step, generate either: an intermediate thought that can allow to reach the answer (Thought:), or the final answer (The final answer is:).} \\
\textbf{\emph{(For one shot example)}}: \\
\texttt{\textbf{\emph{User:}} Which member of the famous Beckham couple has the most Instagram followers?}\\
\texttt{\textbf{\emph{Assistant:}} Intermediate question: Who are the members of the famous Beckham couple? \\
Intermediate answer: The members of the Beckham couple are Victoria Beckham and David Beckham.
Intermediate question: How many followers does Victoria Beckham have on Instagram? \\
Intermediate answer: Victoria Beckham has 32.8M followers on Instagram. \\
Intermediate question: How many followers does David Beckham have on Instagram? \\
Intermediate answer: David Beckham has 87.1M followers on Instagram. \\
So the final answer is: David Beckham}
\textbf{\emph{(For both)}}: \\
\texttt{\textbf{\emph{User:} TASK}}
\end{minipage}}
\caption{Prompts for our closed-book LMs.}
\label{fig:cb_prompt}
\end{figure*}

\begin{figure*}
\colorbox{gray!10}{
\begin{minipage}{15cm}
\texttt{\textbf{\emph{System prompt:}} You are a helpful agent that has to provide answers to complex tasks. \\
Tasks can require accessing the web. To access the web, you are equipped with a search engine. \\
To access the search engine, simply generate an intermediate question. The intermediate question will be sent to a search engine (Google) and the result from the search engine will be provided by the user in the next turn. \\
It is best to access the search engine with simple queries, as they are most likely to return the correct answers. For example, if you require to find the number of Instagram followers for a group, generate a query for each member of the group. \\
When the user message ends with an intermediate question, always generate an intermediate answer. \\
If the search results do not answer your intermediate question, begin the intermediate answer with "Intermediate answer: The intermediate answer cannot be inferred from the search results". \\
In that case, you can use your own knowledge to answer the intermediate question or generate a new intermediate question. \\
You can generate additional intermediate questions and answers until arriving at the final answer. Do not generate the same intermediate question twice. \\
When arriving at the final answer, start your prediction with: "So the final answer is:" followed by the answer to the task. \\
If you are unsure what the final answer is, generate the empty string (""). \\
Keep the answer as concise as possible. The answer should either be: a number, a string, a list of strings, or a list of jsons. \\
The answer should be parsed with the python method: json.loads(input\_str). If no answer is found, generate an empty string. \\
The format will be the search results, followed by the task, followed by the model answers from the previous steps. You are provided with one example. Please use exactly the same format as the example.} \\
\textbf{\emph{(For one shot example)}}: \\
\texttt{\textbf{\emph{User:}} Task: Which member of the famous Beckham couple has the most Instagram followers?}
\texttt{\textbf{\emph{Assistant:}} Intermediate question: Who are the members of the famous Beckham couple?} \\
\texttt{\textbf{\emph{User:}} Context1: Posh and Becks: Posh and Becks is a widely used nickname for the British celebrity supercouple Victoria Beckham (née Adams, 'Posh Spice' of the Spice Girls) and David Beckham (a footballer and former England captain). \\
Task: Which member of the famous Beckham couple has the most Instagram followers? \\
Intermediate question: Who are the members of the famous Beckham couple?} \\
\texttt{\textbf{\emph{Assistant:}} Intermediate answer: The members of the Beckham couple are Victoria Beckham and David Beckham.
Intermediate question:  How many followers does Victoria Beckham have on Instagram?} \\
...\\
\textbf{\emph{(For both)}}: \\
\texttt{\textbf{\emph{User:} CONTEXT \\ TASK \\ MODEL GENERATIONS}}
\end{minipage}}
\caption{Prompts for our retrieval-augmented LMs.}
\label{fig:ralm_prompt}
\end{figure*}

\begin{figure*}
\colorbox{gray!10}{
\begin{minipage}{15cm}
\texttt{\textbf{\emph{System:}} Imagine that you are imitating humans doing web navigation for a task step by step. At each stage, you can see the webpage like humans by a screenshot and know the previous actions before the current step decided by yourself through recorded history. You need to decide on the first following action to take. You can click on an element with the mouse, select an option, type text, press Enter with the keyboard, scroll up and down, go back to the previous page, or go to a different URL (For your understanding, they are like the click(), select\_option(), type(), keyboard.press('Enter'), window.scrollBy(), page.goBack(), page.goto() functions in playwright respectively). One next step means one operation. You are also given the option to go to a search engine (Google) and execute a query in one operation. Unlike humans, for typing (e.g., in text areas, text boxes) and selecting (e.g., from dropdown menus or <select> elements), you should try directly typing the input or selecting the choice, bypassing the need for an initial click. You should not attempt to create accounts, log in or do the final submission. Terminate when you deem the task complete or if it requires potentially harmful actions.You are asked to complete the following task: \textbf{TASK}} 
\end{minipage}}
\caption{System prompt for \seeactexplorer{}.}
\label{fig:spa_sys_prompt}
\end{figure*}

\begin{figure*}
\colorbox{gray!10}{
\begin{minipage}{15cm}
\texttt{\textbf{\emph{User:}} Previous actions: \\
\textbf{PREVIOUS ACTIONS}}\\
\texttt{The screenshot below shows the webpage you see. Follow the following guidance to think step by step before outlining the next action step at the current stage: \\
(Original plan) \\
\textbf{THE ORIGINAL PLAN} \\
(History) \\
\textbf{HISTORY PER STEP}} \\ 
\texttt{(Current Webpage Identification) \\
Firstly, think about what the current webpage is. \\
(Previous Action Analysis) \\
Secondly, combined with the screenshot, analyze each step of the previous action history and their intention one by one. Particularly, pay more attention to the last step, which may be more related to what you should do now as the next step. \\
Specifically, if the last action involved a TYPE, always evaluate whether it necessitates a confirmation step, because typically a single TYPE action does not make effect. (often, simply pressing 'Enter', assuming the default element involved in the last action, unless other clear elements are present for operation). \\
(Screenshot Details Analysis) \\
Closely examine the screenshot to check the status of every part of the webpage to understand what you can operate with and what has been set or completed...} \\
\texttt{(Relevant information) \\
Relevant information that from the webpage to perform the task. Make sure this information can be understood from the webpage. This information will be passed to the next steps. If the webpage does not display any information to perform the task, say "no new infromation". You can use the next steps to find more information or verify information you are unsure of. \\
(Refined plan) \\
A refined plan on how to solve the task that will be passed to next steps. If the original task has been completed, say: "Terminating, the task has been completed". If the task required finding information in the web, add: "Task answer:" followed by the relevant information. Keep the answer as concise as possible. The answer should either be: a number, a string, a list of strings, or a list of jsons. The answer should be parsed with the python method: json.loads(input\_str). If no answer is found, generate an empty string.} \\
\texttt{(Next Action Based on Webpage and Analysis) \\
Then, based on your analysis, in conjunction with human web browsing habits and the logic of web design, decide on the following action. And clearly outline which element in the webpage users will operate with as the first next target element, its detailed location, and the corresponding operation. If you require searching or verifying information you can use the web, for example Google. \\
To be successful, it is important to follow the following rules: \\
1. You should only issue a valid action given the current observation. \\
2. You should only issue one action at a time \\
3. For handling the select dropdown elements on the webpage, it's not necessary for you to provide completely accurate options right now. The full list of options for these elements will be supplied later.}
\end{minipage}}
\caption{Prompt for the first \seeactexplorer{} model call at each step. At the first call, the model describes the necessary action, similarly to \seeact{}, but can also pass information to future steps and revise its execution plan. }
\label{fig:spa_first_turn_prompt}
\end{figure*}

\begin{figure*}
\colorbox{gray!10}{
\begin{minipage}{15cm}
\texttt{\textbf{\emph{Assistant:}}  \\
\textbf{PREVIOUS OUTPUT}}\\
\texttt{\textbf{\emph{User:}} \\
(Reiteration) \\
First, reiterate your next target element, its detailed location, and the corresponding operation. \\
(Multichoice Question) \\
Below is a multi-choice question, where the choices are elements in the webpage. All elements are arranged in the order based on their height on the webpage, from top to bottom (and from left to right). This arrangement can be used to locate them. From the screenshot, find out where and what each one is on the webpage, taking into account both their text content and HTML details. Then, determine whether one matches your target element. Please examine the choices one by one. Choose the matching one. If multiple options match your answer, choose the most likely one by re-examining the screenshot, the choices, and your further reasoning. \\
If none of these elements match your target element, please select AA. None of the other options match the correct element. If you want to scroll up or down the page, select AB. Scroll (up or down). If you want to go a different URL such as Google.com, please select AC. Go to a different URL and pass the full URL as the value. If you want to run a query in a search engine, please select AD. Execute a query in a search engine and pass the query as the value. \\
\textbf{PAGE ELEMENTS} \\
(Final Answer) \\
Finally, conclude your answer using the format below. Ensure your answer is strictly adhering to the format provided below. Please do not leave any explanation in your answers of the final standardized format part, and this final part should be clear and certain. The element choice, action, and value should be in three separate lines. \\
Format: \\
ELEMENT: The uppercase letter of your choice. (No need for PRESS ENTER) \\
ACTION: Choose an action from {CLICK, SELECT, TYPE, GOTO, SEARCH, GOBACK, SCROLL, PRESS ENTER, TERMINATE, NONE}. \\
VALUE: Provide additional input based on ACTION. \\
The VALUE means: \\
If ACTION == TYPE, specify the text to be typed.\\
If Action == GOTO, specify the url that you want to visit. \\ 
If Action == SEACH, specify query you want to be executed. \\ 
If Action == SCROLL, specify if you want to scroll up or down, If ACTION == SELECT, indicate the option to be chosen. Revise the selection value to align with the available options within the element. \\
If ACTION == CLICK, PRESS ENTER, TERMINATE or NONE, write "None".}
\end{minipage}}
\caption{Prompt for the second \seeactexplorer{} model call at each step. At the second call, the model grounds the natural language description to an action with one of the HTML elements.}
\label{fig:spa_second_turn_prompt}
\end{figure*}

\end{document}